# On the Qualitative Comparison of Decisions Having Positive and Negative Features

**Didier Dubois**                                                    DUBOIS@IRIT.FR
**Hélène Fargier**                                                   FARGIER@IRIT.FR
*Université de Toulouse*
*IRIT- CNRS, 118 route de Narbonne*
*31062 Toulouse Cedex, France*

**Jean-François Bonnefon**                                           BONNEFON@UNIV-TLSE2.FR
*Université de Toulouse*
*CLLE (CNRS, UTM, EPHE)*
*Maison de la Recherche, 5 al. Machado*
*31058 Toulouse Cedex 9, France*

## Abstract

Making a decision is often a matter of listing and comparing positive and negative arguments. In such cases, the evaluation scale for decisions should be considered bipolar, that is, negative and positive values should be explicitly distinguished. That is what is done, for example, in Cumulative Prospect Theory. However, contrary to the latter framework that presupposes genuine numerical assessments, human agents often decide on the basis of an ordinal ranking of the pros and the cons, and by focusing on the most salient arguments. In other terms, the decision process is qualitative as well as bipolar. In this article, based on a bipolar extension of possibility theory, we define and axiomatically characterize several decision rules tailored for the joint handling of positive and negative arguments in an ordinal setting. The simplest rules can be viewed as extensions of the maximin and maximax criteria to the bipolar case, and consequently suffer from poor decisive power. More decisive rules that refine the former are also proposed. These refinements agree both with principles of efficiency and with the spirit of order-of-magnitude reasoning, that prevails in qualitative decision theory. The most refined decision rule uses leximin rankings of the pros and the cons, and the ideas of counting arguments of equal strength and cancelling pros by cons. It is shown to come down to a special case of Cumulative Prospect Theory, and to subsume the "Take the Best" heuristic studied by cognitive psychologists.

## 1. Introduction

Not only personal experience, but also psychological experiments suggest that making a decision is often a matter of listing and comparing the positive and negative features of the alternatives (Cacioppo & Berntson, 1994; Osgood, Suci, & Tannenbaum, 1957; Slovic, Finucane, Peters, & MacGregor, 2002). Individuals evaluate alternatives or objects by considering their positive and negative aspects in parallel — for instance, when choosing a movie, the presence of a good actress is a positive argument; a noisy theater or bad critiques are negative arguments. Under this *bipolar* perspective, comparing two decisions amounts to comparing two pairs of sets, that is the sets of pros and cons attached to one decision with the sets of pros and cons attached to the other. Cumulative Prospect Theory (Tversky & Kahneman, 1992) is an explicit attempt at accounting for positive and negative arguments in the numerical setting. It proposes to compute a "net predisposition" of a decision, as the difference between two set functions (capacities) taking values on the positive real line, the first one measuring the importance of the group of positive features, the second one the importance of the group of negative features. More general numerical models, namely bi-capacities (Grabisch & Labreuche, 2005) and bipolar capacities (Greco, Matarazzo, & Slowinski, 2002) encompass more sophisticated situations where criteria are not independent from each other. These numerical approaches to bipolar decision contrast with standard decision theory, which does not account for the bipolarity phenomenon. Indeed, utility functions are defined up to an increasing affine transformation, which does not preserve the value 0.

However, cognitive psychologists have claimed that while arguments featured in a decision process can be of different strengths, decision-makers are likely to consider these degrees of strength at the ordinal level rather than at the cardinal level (Gigerenzer, Todd, & the ABC group, 1999). Individuals appear to consider very few arguments (i.e., the most salient ones) when making their choice, rather than to attempt an exact numerical computation of the merits of each decision (Brandstätter, Gigerenzer, & Hertwig, 2006). In sum, cognitive psychologists have claimed that human decision processes are likely to be largely qualitative as well as bipolar.

The last 10 years have also witnessed the emergence of qualitative decision theory in Artificial Intelligence (Doyle & Thomason, 1999). For instance, qualitative criteria like Wald's rule (Wald, 1950/1971) have been axiomatized along the line of decision theory (Brafman & Tennenholtz, 2000) as well as variants or extensions thereof; see the survey of Dubois and Fargier (2003). So-called conditional preference networks (Boutilier, Brafman, Domshlak, Hoos, & Poole, 2004) have been introduced for an easier representation of preference relations on multidimensional sets of alternatives, using local preference statements interpreted in the ceteris paribus style. This recent emergence of qualitative decision methods in Artificial Intelligence is partly motivated by the traditional stress on qualitative representations in reasoning methods. It is also due to the fact that numerical data are not available in many AI applications, for example because there is no point in requiring very precise evaluations from the user (e.g., for recommender systems).

These models use preference relations which express statements like "decision $a$ is better than decision $b$ for an agent." However, these preference relations cannot deal with bipolarity—more precisely, they cannot express the simple notion that people know what is good and what is bad for them, and that these judgments are orthogonal to judgments about what is "best" in a given situation. Sometimes, the best available choice can be detrimental anyway—and yet, on other occasions, even the worst option is still somewhat desirable. (Note that the "best" or the "worst" option can be the statu quo option, i.e., the option of not making any active choice.) To fully capture these situations, it is necessary to have an absolute landmark or reference point in the model, which expresses neutrality and separates the positive values from the negative values. While emphasizing qualitative approaches to decision and choice, the Artificial Intelligence literature in this area has somewhat neglected the fact that preference orderings are not enough to express the fact that an option can be good or bad per se.

Other qualitative formalisms capable of representing preference exploit value scales that include some reference points, e.g., fuzzy constraint satisfaction problems (Dubois, Fargier,

& Prade, 1996) and possibilistic logic (Benferhat, Dubois, & Prade, 2001). There, the merit of a decision is evaluated on different criteria by means of some kind of utility functions mapping into a bounded ordinal scale whose bottom value expresses an unacceptable degree of violation, and whose top value expresses the absence of a violation. Decisions are then ranked according to the merit of their worst evaluation, following a pessimistic attitude. But this kind of approach is not bipolar, as it only handles negative arguments: the absolute landmark expresses unacceptability, not neutrality. Neutrality is only present by omission, when no constraint is violated, and no decision exists that could be better than neutral.

Another kind of bipolarity is accounted for by Benferhat, Dubois, Kaci, and Prade (2006), who distinguish between prioritized constraints on the one hand, and goals or desires on the other hand. Constraints (expressed as logical formulas) are given a prominent role: they guide the initial selection of the most tolerated decisions. Positive preferences (goals and desires) are then taken in consideration to discriminate among this set of tolerated decisions. As a consequence, positive evaluations (no matter how positive) can never trump negative evaluations (no matter how negative).

Finally, the topic of argumentation in reasoning has gained considerable interest in artificial intelligence in the last ten years or so. This is a natural way of coping with inconsistency in knowledge bases (Besnard & Hunter, 2008). Argumentation is naturally of a bipolar nature, since the construction of arguments consists in collecting reasons for deriving a proposition and reasons for deriving its negation, before proceeding to arbitration between arguments of various weights (Cayrol & Lagasquie-Schiex, 2005). However it is not clear that the strength of an argument in inconsistency-tolerant reasoning should be defined as a number. It sounds more natural to adopt a qualitative approach to the bipolar nature of argument-based reasoning. Moreover AI-based decision procedures also naturally rely on argumentation, so as to facilitate the process of explicating the merits of a decision (Amgoud & Prade, 2004, 2006).

In the present paper,[1] we aim at proposing a bipolar and qualitative setting, equipped with a family of decision rules, in which the decision is based on the comparison of positive and negative arguments whose strength can only be assessed in an ordinal fashion. We insist on the assumption that positive and negative evaluations *share a common scale;* then evaluations having one polarity can trump the other polarity, e.g., a strong positive (resp. negative) argument can win against a weaker negative (resp. positive) argument. Indeed, this is precisely the idea behind the intuitive procedure of weighing the pros against the cons, i.e., finding out the heavier side.This cannot be done without a common importance scale. We also adopt a systematic approach, in which we formalize and axiomatically characterize a set of procedures that are simultaneously ordinal and bipolar.

The paper is structured as follows. Section 2 introduces our framework for qualitative bipolar choice. Then, Section 3 presents two basic qualitative bipolar decision rules. In Section 4, we show how the basic properties of bipolar reasoning can be expressed axiomatically, and, taking one step further, which axioms can capture the principles of *qualitative* bipolar decision-making. Section 5 studies decision rules that are more decisive than the basic rules of Section 3, without giving up their qualitative nature. Finally, Sections 6 and 7 relate some of our rules to a range of other approaches, and identify avenues of future research. Proofs of properties and representation theorems are provided in the appendix.

---

1. This paper is an extended version of the work of Dubois and Fargier (2006). It analyzes additional decision rules, and provides a full axiomatization of the cardinality based rule.

## 2. A Framework for Qualitative Bipolar Decisions

A formal framework for qualitative bipolar multicriteria decision should consist of a finite set $D$ of potential decisions $a, b, c, \ldots$; a set $X$ of criteria or arguments, viewed as attributes ranging on a bipolar scale, say $V$; and a totally ordered scale $L$ expressing the relative importance of criteria or groups of criteria. In this article, we use the simplest possible bipolar scale $V = \{-, 0, +\}$, whose elements reflect negativity, neutrality, and positivity, respectively. With this scale, any argument in $X$ is either completely against, totally irrelevant, or totally in favor of each decision in $D$. But for the focus on bipolarity, this is a simpler approach than many multi-criteria decision-making frameworks where each criterion $x \in X$ is rated on a numerical scale, like multi-attribute utility theory. However, qualitative evaluations are often closer to human capabilities than numerical ones. On this basis it is useful to see how far we can go with a very rough modelling of preference in the bipolar situation. Indeed, if a problem can be solved in a rigorous way without resorting to numerical evaluations, more sophisticated techniques are not needed.

Let $A = \{x, x(a) \neq 0\}$ be the set of relevant arguments for decision $a$. It only contains arguments that matter about $a$, either because they are good things or because they are bad things. Now let $A^- = \{x, x(a) = -\}$ be the set of arguments against decision $a$, and $A^+ = \{x, x(a) = +\}$ the set of arguments in favor of $a$. Considering the sets $A^-$ and $A^+$ amounts to enumerating the pros and the cons of $a$. Thus, comparing decisions $a$ and $b$ amounts to comparing the pairs of disjoint sets $(A^-, A^+)$ and $(B^-, B^+)$. Obviously, if $A^- \subseteq B^-, B^+ \subseteq A^+$, then $a$ should clearly be preferred to $b$ in the wide sense. This is the basic property *bivariate monotony* that any bipolar decision rule should obey.

For the sake of simplicity, we assume in the following that $X$ is divided into two subsets. $X^+$ is the set of positive arguments taking their value in $\{0, +\}$; and $X^-$ is the set of negative arguments taking their value in $\{-, 0\}$. In this simplified model, it is no longer possible to have $A^- \cap B^+ \neq \emptyset$ nor $A^+ \cap B^- \neq \emptyset$. This, however, is done without loss of generality and will not affect the validity of our results in the ordinal setting. Indeed, any $x$ whose evaluation may range in the full domain $\{-, 0, +\}$ can be duplicated, leading to an attribute $x_+$ in $X^+$ and an attribute $x_-$ in $X^-$. Furthermore, this transformation can generalize our framework to arguments that have both a positive and a negative side (e.g., eating chocolate).

The scale $L$ measuring the importance of the arguments has top $1_L$ (full importance) and bottom $0_L$ (no importance). Within a qualitative approach, $L$ can be finite. As it is common with set functions, we make the hypothesis that the importance of a group of arguments only depends on the importance of the individual arguments in the group. With this assumption of independence, levels of importance can be directly attached to the elements of $X$ by a function $\pi : X \mapsto L$, from which the importance of any group of arguments can also be derived. $\pi(x) = 0_L$ means that the decision-maker is indifferent to argument $x$; $\pi(x) = 1_L$ means that the argument possesses the highest level of attraction or repulsion (according to whether it applies to a positive or negative argument). $\pi$ is supposed to be non trivial, that is, at least one $x$ has a positive importance level.

**Example 1** (Luc's Holidays). *Luc will provide us with one of our running examples. He is considering two holiday destinations, and has listed the pros and cons of each. Option* [a] *has scenic landscapes (a strong pro), but it is very expensive, and the local airline has a terrible reputation (two strong cons). Option* [b] *is in a non-democratic region, which Luc considers a strong con. On the other hand, option* [b] *has a tennis court, a disco, and a swimming pool. These are three pros, but not very decisive. They do matter, but not as much as the other arguments. Note that Luc can only give a rough evaluation of how strong a pro or a con is. He can only say that the gorgeous landscapes, the indecent price, the terrible reputation of the airline company, and the non-democratic governance are four arguments of comparable importance; and that the swimming pool, tennis and disco are three arguments of comparable importance, but not as important as the previous ones. Formally, let:*

$$X^+ = \{landscape^{++}, tennis^+, pool^+, disco^+\} \quad \text{be the subset of pros,}$$
$$X^- = \{price^{--}, airline^{--}, governance^{--}\} \quad \text{be the subset of cons.}$$

*Strong arguments are* $landscape^{++}$, $price^{--}$, $airline^{--}$, *and* $governance^{--}$. *The arguments* $tennis^+$, $pool^+$, *and* $disco^+$ *are weaker. Thus, letting* $\lambda > \beta > 0_L$, *we have:*

$$\pi(lanscape^{++}) = \pi(price^{--}) = \pi(airline^{--}) = \pi(governance^{--}) = \lambda$$
$$\pi(tennis^+) = \pi(pool^+) = \pi(disco^+) = \beta.$$

*Finally options* [a] *and* [b] *are described by the following sets of arguments:*

*Options* [a] : $\quad A^+ = \{landscape^{++}\} \qquad\qquad A^- = \{airline^{--}, price^{--}\}$
*Options* [b] : $\quad B^+ = \{tennis^+, pool^+, disco^+\} \quad B^- = \{governance^{--}\}.$

In sum, each attribute $x \in X$ is Boolean (presence vs. absence), but has a *polarity* (its presence is either good or bad, its absence is always neutral), and an *importance* $\pi(x) \in L$. Now, since we are interested in qualitative decision rules, our approach relies on two modelling assumptions:

**Qualitative Scale:** In $L$, there is a big step between one level of merit and the next one. Arguments are ranked in terms of the *order of magnitude* of their figure of importance by means of the mapping $\pi$.

**Focalization:** The order of magnitude of the importance of a group $A$ of arguments with a prescribed polarity is the one of the most important arguments in the group. This assumption suits the use of a qualitative scale, as it means that weak arguments are negligible compared to stronger ones.

Technically, these assumptions suggest the use of the following measure of importance of a set of arguments

$$\mathrm{OM}(A) = \max_{x \in A} \pi(x).$$

In other terms, we simply use qualitative possibility measures (Lewis, 1973; Dubois, 1986) interpreted in term of order of magnitude of importance.

The next sections examine several decision rules relying on the use of $\mathrm{OM}(A)$, that can be defined for balancing pros and cons. We will see that in the Luc example, some will prefer option [a], some will prefer option [b], some will regard the two options as equally attractive, and some will find it impossible to make a decision. We begin in Section 3 with two basic rules that only take into account the most important arguments. The corresponding ordering can be complete or partial, and is usually weakly discriminant. This is due to the immediate use of the order-of-magnitude evaluations, that leads to very rough decision rules. Refinements of these basic rules will be proposed in Section 5. They apply elementary principles of simplification (discarding arguments that are relevant to both decisions, sometimes cancelling opposite arguments of the same strength) before making a choice. Most of these refinements thus obey a form of preferential independence.

## 3. Elementary Qualitative Bipolar Decision Rules

The two elementary decision rules in this Section differ by one basic feature: the first one treats positive and negative arguments separately; the second one allows for a comparison of the relative strengths of positive and negative arguments, one side possibly overriding the other.

Each decision rule defines a preference relation. Since the relations presented here are not necessarily complete nor transitive, let us recall some definitions, prior to presenting these two decision rules.

**Definition 1.** *For any relation $\succeq$, one can define:*
- *Its symmetric part:*         $A \sim B \iff A \succeq B \text{ and } B \succeq A$
- *Its asymmetric part:*        $A \succ B \iff A \succeq B \text{ and } not(B \succeq A)$
- *An incomparability relation:*   $A \diamond B \iff not(A \succeq B) \text{ and } not(B \succeq A)$

$\succeq$ is said to be *quasi-transitive* when $\succ$ is transitive. The transitivity of $\succeq$ obviously implies its quasi-transitivity, whether or not it is complete. The converse implication generally does not hold. When the relation is complete, $\diamond$ is empty. $\succeq$ is said to be a *weak order* if and only if it is complete (and thus reflexive) and transitive.

In the following, we also use the notion of refinement of a relation:

**Definition 2.** $\succeq'$ *refines* $\succeq$ *if and only if* $\forall A, B : A \succ B \Rightarrow A \succ' B$

The refined relation $\succeq'$ thus follows the strict preference of $\succeq$ when any, but can also make a difference between decisions in case $\succeq$ cannot—that is, it may happen that $A \succ' B$ while $A \sim B$ or $A \diamond B$.

### 3.1 A Bipolar Qualitative Pareto Dominance Rule

The order of magnitude of a bipolar set $A$ is no longer a unique value in $L$ like in the unipolar case, but a pair $(\text{OM}(A^+), \text{OM}(A^-))$. Pairs and more generally vectors of evaluations can be easily compared to others using the classical principle of Pareto comparison. This yields the following rule, which does not assume commensurateness between the evaluations of positive and negative arguments:

**Definition 3.** $A \succeq^{\text{Pareto}} B \iff OM(A^+) \geq OM(B^+) \text{ and } OM(A^-) \leq OM(B^-).$

What would $\succ^{\text{Pareto}}$ conclude on Luc's example? Luc has a strong argument for option [a], but only weak arguments for option [b] : $\text{OM}(A^+) > \text{OM}(B^+)$. In parallel, Luc has strong arguments both against option [a] and against option [b]: $\text{OM}(A^-) = \text{OM}(B^-)$. As a consequence, $A \succ^{\text{Pareto}} B$, and Luc will choose option [a].

Let us lay bare in more details the cases where $A \sim^{\text{Pareto}} B$, $A \succ^{\text{Pareto}} B$, $A \bowtie^{\text{Pareto}} B$.

- $A$ and $B$ are indifferent if and only if their salient positive aspects as well as their salient negative aspects share the same order of magnitude, i.e., $\text{OM}(A^+) = \text{OM}(B^+)$ and $\text{OM}(A^-) = \text{OM}(B^-)$;

- $B$ is negligible compared to $A$ ($A \succ^{\text{Pareto}} B$) in two cases: either $\text{OM}(A^+) \geq \text{OM}(B^+)$ and $\text{OM}(A^-) < \text{OM}(B^-)$, or $\text{OM}(A^+) > \text{OM}(B^+)$ and $\text{OM}(A^-) \leq \text{OM}(B^-)$. The case $B \succ^{\text{Pareto}} A$ can be described symmetrically.

- In other cases, there is a conflict and $A$ is not comparable with $B$ ($A \bowtie^{\text{Pareto}} B$).

$\succeq^{\text{Pareto}}$ is obviously reflexive and transitive. It collapses to Wald's pessimistic ordering (Wald, 1950/1971) when $X = X^-$, and to its optimistic max-based counterpart when $X = X^+$. Note that $\succeq^{\text{Pareto}}$ is partial — and maybe too partial. For example, when a decision has both pros and cons, $\succeq^{\text{Pareto}}$ concludes that it is incomparable to a decision that has not any pro nor con. This can be quite counter-intuitive, as shown in the following example.

**Example 2** (Lucy and the Riviera estate). *Being short of money, Lucy was planning to spend the summer at home. But she is now offered to spend part of the summer at her brother's paradisiacal estate on the Riviera. The only inconvenient of this arrangement is that Lucy finds her sister-in-law mildly annoying.*

*Formally, let: $X = \{estate^{++}, inlaw^-\}$, with $\pi(estate^{++}) > \pi(inlaw^-)$.*
*The two options, going to the riviera or staying home are described as follows:*
 *Option [a]:* $A^+ = \{estate^{++}\}$ *and* $A^- = \{inlaw^-\}$.
 *Option [b]:* $B^+ = \{\}$   $B^- = \{\}$

The common intuition about the Lucy case is that if really her sister-in-law is only mildly annoying, and if the estate is so fantastic, Lucy is likely to prefer to go there rather than stay at home. However, $\succeq^{\text{Pareto}}$ cannot predict this preference because it finds the two options incomparable: $\text{OM}(A^+) > \text{OM}(B^+)$, but $\text{OM}(A^-) > \text{OM}(B^-)$.

Another drawback of $\succeq^{\text{Pareto}}$ becomes clear when the two decisions have the same order of magnitude on one of the two dimensions:

**Example 3** (Luka and the gyms). *Luka is considering buying membership in one of two gyms. Option [a] is very expensive, which is a strong con. Option [b] is also very expensive, but it comes with the small bonus of having a squash court (this is a small bonus because Luka is not sure yet he will want to use that court). On the other hand, option [b] also has a drawback of medium importance, that is, it is inconveniently located. Location is not as important as price in Luka's mind, but it is still more important than the presence of a squash court. Formally:*
 $X = \{squash^+, location^{--}, price^{---}\}$   *with*
 $\pi(squash^+) < \pi(location^{--}) < \pi(price^{---})$.

*The options are described as follows:*
   *Option* [a]:   $A^+ = \{squash^+\}$    $A^- = \{location^{--}, price^{---}\}$,
   *Option* [b]:   $B^+ = \emptyset$           $B^- = \{price^{---}\}$.

It follows from Definition 3 that Luka will prefer option [a], because $OM(A^+) > OM(B^+)$ whilst $OM(A^-) = OM(B^-)$. This is intuitively unsatisfying, however, for we would expect Luka to examine more carefully the fact that option [a] is inconveniently located, which is a moderately strong con, rather than to decide on the basis of a very weak argument, that is, the squash court. In other terms, $\succeq^{\text{Pareto}}$ does not completely obey the principle of Focalization discussed in the introduction: An argument of a lower level (the squash court) can determine a choice even though an argument of a higher level (the location) would have pointed at the opposite direction.

This problem with $\succeq^{\text{Pareto}}$ is partly rooted in the fact that it does not capture the assumption that positive and negative evaluations *share a common scale*. The fact that an argument may be stronger than an argument of the opposite polarity is never taken into account. We will now propose a more realistic rule that captures this assumption.

## 3.2 The Bipolar Possibility Relation

In this section, we propose a decision rule for comparing $A$ and $B$ that focuses on arguments of maximal strength in $A \cup B$, i.e., those at level $\delta = \max_{y \in A \cup B} \pi(y) = OM(A \cup B)$. The principle underlying this rule is simple: any argument against $A$ (resp. against $B$) is an argument pro $B$ (resp. pro $A$), and conversely. The most supported decision is then preferred.

**Definition 4.** $A \succeq^{\text{BiPoss}} B \iff OM(A^+ \cup B^-) \geq OM(B^+ \cup A^-)$.

This rule decides that $A$ is at least as good as $B$ iff, at the highest level of importance, there are arguments in favor of $A$ or arguments attacking $B$. Clearly $A \succ^{\text{BiPoss}} B$ iff, at the highest level, there is at least a positive element for $A$ or an element against $B$, but no element against $A$ and no element pro $B$. Obviously, $\succeq^{\text{BiPoss}}$ collapses to Wald's pessimistic ordering if $X = X^-$, and to its optimistic counterpart when $X = X^+$. In some sense, this comparison yields the most straightforward way of generalizing possibility orderings to the bipolar case.

**Proposition 1.** $\succeq^{\text{BiPoss}}$ *is complete and quasi-transitive.*

In other terms, the strict part of the relation $\succeq^{\text{BiPoss}}$ is transitive, but the associated indifference relation is generally not: $A \sim^{\text{BiPoss}} B$ and $B \sim^{\text{BiPoss}} C$ do not imply $A \sim^{\text{BiPoss}} C$. For instance, let us denote $a^+ = OM(A^+)$, $a^- = OM(A^-)$, $b^+ = OM(B^+)$, $b^- = OM(B^-)$, $c^+ = OM(C^+)$, and $c^- = OM(C^-)$. Assume $\max(a^+, b^-) = \max(a^-, b^+)$ and $\max(b^+, c^-) = \max(b^-, c^+)$. Assume $b^+ = b^- = 1_L$. Then the two equalities hold regardless of the values $a^+, a^-, c^+, c^-$. So the values $\max(a^+, c^-)$ and $\max(a^-, c^+)$ can be anything. Similarly $A \succ^{\text{BiPoss}} B$ and $B \sim^{\text{BiPoss}} C$ do not imply $A \succ^{\text{BiPoss}} C$ – counter examples can be built setting $c^+ = c^- = 1_L$.

In the case of Luc (Example 1), $A = \{landscape^{++}, airline^{--}, price^{--}\}$, and $B = \{governance^{--}, tennis^+, swimming^+, disco^+\}$. From the perspective of $\succeq^{\text{BiPoss}}$, both options are equivalently bad, since $OM(A^+ \cup B^-) = OM(B^+ \cup A^-)$. Likewise, in the case of Luka (Example 3), $\succeq^{\mathrm{BiPoss}}$ will regard the two gyms as equivalently bad because of their high price. Now, in the case of Lucy (Example 2), remember that $\succeq^{\mathrm{Pareto}}$ regarded options [a] and [b] as incomparable, where $A = \{estate^{++}, inlaws^-\}$ and $B$ is the empty set. In contrast, $\succeq^{\mathrm{BiPoss}}$ will consider, in line with common intuition, that Lucy will prefer to go to the Riviera estate, since $\mathrm{OM}(A^+ \cup B^-) > \mathrm{OM}(B^+ \cup A^-)$.

$\succeq^{\mathrm{BiPoss}}$ is very different and arguably less dubious than $\succeq^{\mathrm{Pareto}}$. But, as shown by Luc and Lukas's examples, it is a very rough rule that may be not decisive enough. This weakness of $\succeq^{\mathrm{BiPoss}}$ is rooted in the usual drowning effect of possibility theory: when an argument of high importance is attached to both decisions (e.g., the ludicrous price for the two gyms), it will trump all arguments of lesser importance (e.g., the squash court, but also the location).[2] Variants of $\succeq^{\mathrm{BiPoss}}$ will be presented in Section 5 that overcome this difficulty. Nevertheless, the rule $\succeq^{\mathrm{BiPoss}}$ alone has the merit of capturing the essence of ordinal decision-making, as shown by the axiomatic study presented in the next section.

## 4. Axioms for Ordinal Comparison on a Bipolar Scale

In the previous sections, we have proposed a framework and some decision rules that intend to capture the essence of qualitative bipolar decision-making. In the present section, we adopt the opposite strategy, that is, we formalize natural properties that such a qualitative bipolar preference relation should obey—and we show that our framework is not only sound (it obeys the aforementioned properties) but also complete. Our main result is a representation theorem stating that any preference relation that satisfies these properties is equivalent to $\succeq^{\mathrm{BiPoss}}$.

Let $\succeq$ be an abstract preference relation on $2^X$, $A \succeq B$ meaning that decision $A$ is at least as good as decision $B$. As $\succeq$ compares sets of decisions, we call it a *set-relation*. First of all, we introduce general properties (e.g., reflexivity or monotony) $\succeq$ should sensibly obey to be a well-behaved bipolar set-relation, be it qualitative or not. Then, we introduce axioms that characterize *qualitative* bipolar set-relations.

### 4.1 Axioms for Monotonic Bipolar Set-Relations

First of all, as for any preference relation, we shall assume minimal working conditions for a sensible framework, such as reflexivity (R) and quasi-transitivity (QT).[3]

The basic notion of bipolar reasoning over sets of arguments is the separation of $X$ into good and bad arguments. The first axiom thus states that any argument is either positive or negative in the wide sense, i.e., either not worse than or nor better worse than nothing:

**Clarity of Arguments (CA)** $\forall x \in X, \{x\} \succeq \emptyset$ or $\emptyset \succeq \{x\}$.

---

2. The drowning effect is also at work in $\succeq^{\mathrm{Pareto}}$, within the comparison of $A^+$ and $B^+$, or within the comparison of $A^-$ and $B^-$. Ceteris paribus, a destination with gorgeous landscapes *plus* a swimming pool is not preferred to a destination with gorgeous landscapes and no pool.

3. Although weak, these assumptions are relaxed in some relational approaches to multicriteria evaluation, where the aggregation process produces cycles in the preference relation. Nevertheless, in such methods, even when the resulting relation is not transitive, the next step is to build a transitive approximation to it.

One can then partition $X$, differentiating positive, negative and null arguments:

$$X^+ = \{x, \{x\} \succ \emptyset\}$$
$$X^- = \{x, \emptyset \succ \{x\}\}$$
$$X^0 = \{x, \emptyset \sim \{x\}\}$$

Now, arguments to which the decision-maker is indifferent should obviously not affect his or her preference. This is the meaning of the next axiom, which allows to forget about $X^0$ without loss of generality:

**Status Quo Consistency (SQC)**
If $\{x\} \sim \emptyset$ then $\forall A, B : A \succeq B \;\; \Leftrightarrow \;\; A \cup \{x\} \succeq B \;\; \Leftrightarrow \;\; A \succeq B \cup \{x\}$ .

Let us now discuss the property of monotony. Monotony in the sense of inclusion ($A \subseteq B \implies B \succeq A$) can obviously not be obeyed as such in a bipolar framework. Indeed, if $B$ is a set of negative arguments, it generally holds that $A \succ A \cup B$. We rather need axioms of monotony *specific to* positive and negative arguments—basically, the one of bipolar capacities (Greco et al., 2002), expressed in a comparative way.

**Positive Monotony** $\quad \forall C, C' \subseteq X^+, \forall A, B : \quad A \succeq B \quad \Rightarrow \quad C \cup A \quad \succeq \quad B \setminus C'.$
**Negative Monotony** $\quad \forall C, C' \subseteq X^-, \forall A, B : \quad A \succeq B \quad \Rightarrow \quad C \setminus A \quad \succeq \quad B \cup C'.$

We will see that the bivariate monotony property is captured by this pair of axioms.

Now, another assumption is that only the positive side and the negative side of $A$ and $B$ are to be taken into account when comparing them: if $A$ is at least as good as $B$ on both the positive and the negative sides, then $A$ is at least as good as $B$. This is expressed by the axiom of weak unanimity.

**Weak Unanimity** $\forall A, B, A^+ \succeq B^+$ and $A^- \succeq B^- \Rightarrow A \succeq B$.

The set-relations presented in the previous Section obviously satisfy weak unanimity. Finally, we add a classical axiom of non triviality:

**Non-Triviality**: $X^+ \succ X^-$.

It leads to the following generalization of comparative capacities:

**Definition 5.** *A relation on a power set $2^X$ is a* monotonic bipolar set-relation *if and only if it is reflexive, quasi-transitive and satisfies the properties CA, SQC, Non-Triviality, Weak unanimity, Positive and Negative Monotony.*

As expected, a monotonic bipolar set-relation safisfies the bivariate monotony property: using the above conventions for positive and negative arguments in subsets $A$ and $B$, if $B^+ \subseteq A^+$ (resp. $A^- \subseteq B^-$), then using Clarity of Arguments and Positive (resp. Negative) Monotony, it follows that $A^+ \succeq B^+$ (resp. $A^- \succeq B^-$), hence $A \succeq B$ due to weak unanimity.

**Proposition 2.** $\succeq^{\text{BiPoss}}$ *is a monotonic bipolar set-relation.*

In the present work, we are interested in set-relations that are entirely determined by the strength and the polarity of the individual arguments in $X$. We denote by $\mathbb{X} = X \cup \{0\}$ the set of individual arguments in $X$, adding an element $0$ so as to keep track of the polarity of the arguments. The basic information about arguments is captured by the restriction of $\succeq$ to $\mathbb{X}$. Formally it is defined by:

$$x \succeq_{\mathbb{X}} y \iff \{x\} \succeq \{y\}$$
$$x \succeq_{\mathbb{X}} 0 \iff \{x\} \succeq \emptyset$$
$$0 \succeq_{\mathbb{X}} x \iff \emptyset \succeq \{x\}$$

From now on, we will call $\succeq_{\mathbb{X}}$ the *ground relation* of $\succeq$.

In agreement with the existence of a totally ordered scale for weighting arguments, the ground relation $\succeq_{\mathbb{X}}$ is supposed to be a weak order. As a consequence, a minimal condition of coherence for $\succeq_{\mathbb{X}}$ is that a preference cannot be reversed when an argument in the preferred set (resp., the least preferred set) is replaced by an even better one (resp., a worse one). This can be viewed as a condition of monotony with respect to $\succeq_{\mathbb{X}}$:

**Monotony w.r.t. $\succeq_{\mathbb{X}}$ or "$\mathbb{X}$-monotony"**
$\forall A, B, x, x'$ such that $A \cap \{x, x'\} = \emptyset$ and $x' \succeq_{\mathbb{X}} x$:

$$A \cup \{x\} \succ B \quad \Rightarrow \quad A \cup \{x'\} \succ B$$
$$A \cup \{x\} \sim B \quad \Rightarrow \quad A \cup \{x'\} \succeq B$$
$$B \succ A \cup \{x'\} \quad \Rightarrow \quad B \succ A \cup \{x\}$$
$$B \sim A \cup \{x'\} \quad \Rightarrow \quad B \succeq A \cup \{x\}$$

This very natural axiom is richer than it seems. For example, it implies a property of substitutability of equally strong arguments of the same polarity—a kind of property that is often called "anonymity" in social choice and decision theory. A kindred property to anonymity should also be required, for positive arguments to block negative arguments of the same strength. This blocking effect should not depend on the arguments themselves, but only on their position on the scale. Hence the axioms of positive and negative cancellation:

**Positive Cancellation(POSC)**
$\forall x, z \in X^+, y \in X^-, \{x, y\} \sim \emptyset$ and $\{z, y\} \sim \emptyset \Rightarrow x \sim_{\mathbb{X}} z$.

**Negative Cancellation (NEGC)**
$\forall x, z \in X^-, y \in X^+, \{x, y\} \sim \emptyset$ and $\{z, y\} \sim \emptyset \Rightarrow x \sim_{\mathbb{X}} z$.

It makes sense to summarize the above requirements into a single axiom, that we call *Simple Grounding:*

**Simple Grounding**
A bipolar set-relation $\succeq$ is said to be simply grounded if and only if $\succeq_{\mathbb{X}}$ is a weak order, $\succeq$ is monotonic with respect to $\succeq_{\mathbb{X}}$ and satisfies positive and negative cancellation.

**Proposition 3.** *The set-relation $\succeq^{\mathrm{BiPoss}}$ is simply grounded.*[4]

## 4.2 Axiomatizing Qualitative Bipolar Set-Relations

Our definition of a monotonic bipolar set-relation (Definition 5) is very general and encompasses numerous models, be they qualitative (e.g., the two rules in section 3) or not (e.g., cumulative prospect theory in its full generality). As we are interested in preference rules that derive from the principles of ordinal reasoning only, we now focus on axioms that account for ordinality.

The ordinal comparison of sets has been extensively used, especially in Artificial Intelligence. The basic principle in qualitative reasoning is *Negligibility*, which assumes that each level of importance can be interpreted as an order of magnitude, much higher than the next lower level.

### Negligibility (NEG)
$\forall A, B, C \subseteq X^+ : A \succ B$ and $A \succ C \Rightarrow A \succ B \cup C$.

Axiom NEG has already been featured around in AI, directly under this form or through more demanding versions. Let us mention the "union property" of nonmonotonic reasoning, or Halpern's (1997) "Qualitativeness" axioms (see Dubois & Fargier, 2004, for a discussion). Lehmann (1996) introduced an axiom of negligibility inside Savage decision theory axiomatics.

Qualitative reasoning generally also comes along with a notion of closeness preservation (which does away with the notion of counting):

### Closeness Preservation (CLO)
$\forall A, B, C \subseteq X^+ : \quad A \sim B$ and $A \sim C \Rightarrow A \sim B \cup C$
$\forall B, C \subseteq X^+ : \quad\quad B \succeq C \Rightarrow B \sim B \cup C$

These axioms were proposed and justified in ordinal reasoning when only one scale needs to be considered (namely, the positive one). But they are not sufficient when a negative scale also needs to be taken into account. We need for example to express that if there is a very bad consequence $B$, so bad that $A \succ B$ and $C \succ B$, then whatever the negative arguments in $A$ and $C$, $B$ is still worse than $A \cup C$:

$$\forall A, B, C : A \succ B \text{ and } C \succ B \Rightarrow A \cup C \succ B.$$

This property is meaningful for negative sets of arguments, and trivial on $X^+$; hence it can be introduced soundly in the framework.

Other cases where sets with both negative and positive elements are compared should also be encompassed. For example, if $A$ is so good that it can cope with globally negative $B$ and also win the comparison with $C$, then $A \cup B$ is still better than $C$:

$$\forall A, B, C : A \cup B \succ \emptyset \text{ and } A \succ C \Rightarrow A \cup B \succ C.$$

---

4. Technically, $\succeq^{\mathrm{Pareto}}$ is also a monotonic bipolar simply grounded set-relation. But this result is quite irrelevant, since this rule never compares the strengths of positive and negative arguments. In other terms, it never happens that arguments of opposite polarities cancel each other, and the condition parts of POSC and NEGC are thus never fulfilled.

And similarly, if a globally negative $A$ $(A \prec \emptyset)$ is so bad that it is outperformed by $C$ $(C \succ A)$ and cannot be enhanced by $B$ $(\emptyset \succ A \cup B)$, then $C \succ A \cup B$, i.e.:

$$\forall A, B, C : \emptyset \succ A \cup B \text{ and } C \succ A \Rightarrow C \succ A \cup B.$$

All these properties can be expressed in the following axiom of global negligibility:

**Global Negligibility (GNEG)**
$\forall A, B, C, D : A \succ B$ and $C \succ D \Rightarrow A \cup C \succ B \cup D$

This is a classic property for purely positive qualitative scales—in that case, it is a consequence of NEG and positive monotony. That is why it is usually not explicitly required in positive frameworks. But when a framework with a positive and a negative scale is needed, the NEG condition is no longer sufficient for getting GNEG. So, in order to keep the property that is a foundation of pure order-of-magnitude reasoning, bipolar qualitative frameworks must explicitly require GNEG.

A similar argument applies to axiom CLO. We must explicitly require a property that is more general than that which is usual for unipolar qualitative scales:

**Global Closeness Preservation (GCLO)**
$\forall A, B, C, D : A \succeq B$ and $C \succeq D \Rightarrow A \cup C \succeq B \cup D$

**Proposition 4.** *The set-relation* $\succeq^{\text{BiPoss}}$ *satisfies GNEG and GCLO.*

Propositions 2, 3 and 4 show that the bipolar possibility relation is a simply grounded monotonic bipolar set-relation satisfying GNEG and GCLO. Applying the principles of qualitative bipolar reasoning described by the previous axioms can also lead to many different but more or less intuitive qualititative rules, for instance the Pareto rule (see Dubois and Fargier (2006) for a full characterization of this rule). But if we are looking for a simple complete decision rule, these axioms provide a full characterization of the Biposs preference relation. Note that the strict part of this rule is governed by axiom GNEG. The indifference part includes pure case of indifference but also cases when one would expect incomparability between decisions rather than indifference proper. This is the most debatable part of the Biposs rule, that will be refined in the sequel.

**Theorem 1.** *The following propositions are equivalent:*

1. $\succeq$ *is a simply grounded* complete *monotonic bipolar set-relation on* $2^X$ *that satisfies GNEG and GCLO.*

2. *There exists a mapping* $\pi : X \mapsto [0_L, 1_L]$ *such that* $\succeq \ \equiv \ \succeq^{\text{BiPoss}}$.

The detailed proof of Theorem 1 is provided in Appendix A. In short, we show that, when $\succeq$ is complete and simply grounded, a ranking $\geq$ can be built that ranks the arguments with respect to their strength. Within $X^+$ and within $X^-$, $\geq$ simply obeys the information captured in $\succeq_{\mathbb{X}}$:

- $\forall x, y \in X^+ \cup X^0, x \geq y \iff x \succeq_{\mathbb{X}} y$;

- $\forall x, y \in X^- \cup X^0, x \geq y \iff x \preceq_{\mathbb{X}} y$.

The relative strength of elements of different signs is deduced from blocking effects. Indeed, when $\{x, y\}$ is preferred to the empty set, the positive argument must be stronger than the negative one (and symmetrically). The two arguments are of equivalent strength when none of them can win: $\{x, y\} \sim \emptyset$ . Formally:

- $\forall x \in X^+, y \in X^-, x \geq y \iff \{x, y\} \succeq \emptyset$;

- $\forall x \in X^+, y \in X^-, y \geq x \iff \emptyset \succeq \{x, y\}$.

The condition of simple grounding then ensures that $\geq$ is a weak order. It can thus be encoded by a mapping $\pi : X \mapsto [0_L, 1_L]$ such that $\pi(x) = 0_L \iff x \in X^0$. Note that this construction is valid for any simple grounded complete bipolar relation, and not only for qualitative ones. The sequel of the proof then uses the axioms for closeness and negligibility to show that $\succeq \equiv \succeq^{\text{BiPoss}}$.

### 4.3 The Principles of Efficiency and Preferential Independence

In summary, the previous Section has shown that $\succeq^{\text{BiPoss}}$ is a natural model of preferences based on bipolar orders of magnitude. In particular, any rule in accordance with GNEG has to follow the strict preference prescribed by $\succ^{\text{BiPoss}}$.

Nonetheless, we have seen in Section 3.2 that $\succeq^{\text{BiPoss}}$ suffers from a drowning effect, as usual in standard possibility theory. For instance, when $B$ is included in $A$ and even if all their elements are positive, then $A$ is not necessarily strictly preferred to $B$. This problem is rooted into the fact that CLO concludes to indifference even in some cases where we would like to appeal to the so-called "principle of efficiency" to make the decision. Just like the monotony principle, this axiom is well known on positive sets. Its proper extension to the bipolar framework obviously has one positive and one negative side:

**Positive efficiency** $B \subseteq A$ and $A \setminus B \succ \emptyset \Rightarrow A \succ B$

**Negative efficiency** $B \subseteq A$ and $A \setminus B \prec \emptyset \Rightarrow A \prec B$

The set-relations $\succeq^{\text{BiPoss}}$ and $\succeq^{\text{Pareto}}$ also fail to obey the classical condition of preferential independence, also called the principle of additivity. This condition simply states that arguments present in both $A$ and $B$ should not influence the decision:

**Preferential Independence**: $\forall A, B, C, (A \cup B) \cap C = \emptyset : A \succeq B \iff A \cup C \succeq B \cup C$

This axiom is well known in uncertain reasoning, as one of the fundamental axioms of comparative probabilities (see Fishburn, 1986, for a survey). Note that it implies the above conditions of efficiency (provided that completeness holds).

Except in very special cases where all arguments are of different levels of importance (when $\succeq_{\mathbb{X}}$ is a linear order), these new axioms are incompatible with axioms of ordinality when completeness or transitivity are enforced. It is already true in the purely positive case, i.e. when $X^-$ is empty (Fargier & Sabbadin, 2005). But this impossibility result is not insuperable, as shown in the next Section.

## 5. Refining the Basic Order-of-Magnitude Comparison

In order to overcome the lack of decisiveness of $\succeq^{\text{BiPoss}}$ we can propose comparison principles that refine it, that is, more decisive set-relations $\succeq$ that are still compatible with $\succeq^{\text{BiPoss}}$, such that $A \succ^{\text{BiPoss}} B \Rightarrow A \succ B$. In the following, we shall not consider refining $\succeq^{\text{Pareto}}$ because of its important drawbacks.

### 5.1 The Implicative Bipolar Decision Rule

The implicative decision rule (Dubois & Fargier, 2005) follows the basic focalization principle of $\succeq^{\text{BiPoss}}$: When comparing $A$ and $B$, it focuses on arguments of maximal strength $\text{OM}(A \cup B) = \max_{x \in A \cup B} \pi(x)$ in $A \cup B$. It adds to this principle the following very simple existential principle: $A$ is at least as good as $B$ iff, at level $s$ the existence of arguments in favor of $B$ is counterbalanced by the existence of arguments in favor of $A$, and the existence of arguments against $A$ is counterbalanced by the existence of arguments against $B$. Formally, the implicative bipolar rule can be described as follows:

**Definition 6.** *Let* $\delta = \max_{x \in A \cup B} \pi(x)$;

$$A \succeq^{\text{Impl}} B \iff and \quad \begin{array}{l} OM(B^+) = \delta \implies OM(A^+) = \delta \\ OM(A^-) = \delta \implies OM(B^-) = \delta \end{array}$$

Let us go back to the examples of Luc, Lucy, and Luka. In the Luc case (Example 1), where $\succeq^{\text{BiPoss}}$ concluded to indifference, $\succeq^{\text{Impl}}$ will rather select Option [a], because there are important arguments against both decisions, whilst only Option [a] is supported by an important pro (there is no important pro supporting option [b]). In the Lucy case (Example 2), $\succeq^{\text{Impl}}$ will follow the strict preference of $\succeq^{\text{BiPoss}}$ and send Lucy to the Riviera. Finally, in the Luka case (Example 3), where the highest level of argument importance features only one con and no pro, on both sides, $\succeq^{\text{Impl}}$ will opine with $\succeq^{\text{BiPoss}}$ and conclude to indifference.

Let us lay bare the cases where $A \succeq^{\text{Impl}} B$. Once again, let us denote $a^+ = \text{OM}(A^+)$, $a^- = \text{OM}(A^-)$, $b^+ = \text{OM}(B^+)$, $b^- = \text{OM}(B^-)$. By definition, $A \succeq^{\text{Impl}} B$ in any of the four following situations:

1. $a^+ = b^+ = a^- = b^-$;

2. $a^+ = b^+ \geq \max(a^-, b^-) > \min(a^-, b^-)$;

3. $a^- = b^- \geq \max(a^+, b^+) > \min(a^+, b^+)$;

4. $\max(a^+, b^-) > \max(a^-, b^+)$.

We thus get the following decomposition of the $\succeq^{\text{Impl}}$ rule:

**Proposition 5.**

- $A \sim^{Impl} B \iff$ *either* $a^+ = b^+ = a^- = b^-$, *or* $a^+ = b^+ > \max(a^-, b^-)$, *or yet* $a^- = b^- > \max(a^+, b^+)$.

- $A \bowtie^{Impl} B \iff$ *either* $a^+ = a^- > \max(b^-, b^+)$, *or* $b^+ = b^- > \max(a^-, a^+)$.

- $A \succ^{Impl} B \iff$ *either* $\max(a^+, b^-) > \max(a^-, b^+)$, *or* $a^+ = a^- = b^- > b^+$, *or yet* $b^+ = b^- = a^+ > a^-$.

It is then easy to check that $\succeq^{\mathrm{Impl}}$ is a bipolar monotonic set-relation. Like $\succeq^{\mathrm{BiPoss}}$, when there are positive arguments only, the set-relation $\succeq^{\mathrm{Impl}}$ collapses to the max rule. It also obeys the principle of weak unanimity. Moreover:

**Proposition 6.** *The set-relation* $\succeq^{\mathrm{Impl}}$ *is transitive.*

Since $\max(a^+, b^-) > \max(a^-, b^+)$, i.e., $A \succ^{\mathrm{Biposs}} B$, is one of the conditions for $A \succ^{\mathrm{Impl}} B$, it obviously follows that:

**Proposition 7.** *Relation* $\succeq^{\mathrm{Impl}}$ *is a refinement of* $\succeq^{\mathrm{BiPoss}}$.

Finally, the situation of incomparability $A \bowtie^{\mathrm{Impl}} B$ arises in two cases only, when $a^+ = a^- > \max(b^-, b^+)$, or in the symmetric case $b^+ = b^- > \max(a^-, a^+)$. In other terms, *incomparability* occurs when one of the two sets displays an *internal contradiction* at the highest level, while the arguments in the other set are too weak to matter. In particular, $a^+ = a^- > 0_L$ if and only if $A \bowtie \emptyset$. For instance, a dangerous travel in an exceptional, and mysterious part of a far tropical forest displays such an internal conflict, and I do not know whether I prefer to stay at home or not. Such a conflict appears also in Luc's first option (which is very attractive but high priced). On the other hand, considering a non conflicting (and non-empty set) $A$, either $\mathrm{OM}(A^+) > \mathrm{OM}(A^-)$ and then $A \succ^{Impl} \emptyset$ (well, the travel is reasonably dangerous, so I prefer to go), or $\mathrm{OM}(A^-) > \mathrm{OM}(A^+)$ and then $\emptyset \succ^{Impl} A$ (there is a war near the border and I prefer to stay at home); in these two latter cases, the Pareto rule would have concluded to an incomparability. The range of incompleteness of $\succeq^{Impl}$ is thus very different from the one of $\succeq^{\mathrm{Pareto}}$, which does not account for any notion of internal conflict.

The $\succeq^{\mathrm{Impl}}$ rule is very interesting from a theoretic descriptive point of view, both because of the way it handles the conflict, and because of the fact that it refines $\succeq^{\mathrm{BiPoss}}$. However, $\succeq^{\mathrm{Impl}}$ is not decisive enough to fully overcome the drowning effect: only the most salient arguments are taken into account. For instance, a very expensive hotel *without* swimming pool is undistinguishable from a very expensive hotel that does include a swimming pool.

Even if it is more decisive than $\succeq^{\mathrm{BiPoss}}$ the $\succeq^{\mathrm{Impl}}$ rule does not satisfy Preferential Independence: like the previous rules, it collapses with Wald's criterion (resp., the max rule) on the negative (resp., positive) sub-scale $X^-$ (resp., $X^+$), and can thus suffer from the drowning effect. To solve this problem, we now leave this family of set-relations and focus on a set of refinements that do satisfy Preferential Independence, and are thus efficient both positively and negatively.

### 5.2 Efficient Refinements of $\succeq^{\mathrm{BiPoss}}$

The following so-called "Discri" rule adds the principle of preferential independence to the ones proposed by $\succeq^{\mathrm{BiPoss}}$, cancelling the elements that appear in both sets before applying the $\succeq^{\mathrm{BiPoss}}$ rule:

**Definition 7.** $A \succeq^{\mathrm{Discri}} B \iff A \setminus B \succeq^{\mathrm{BiPoss}} B \setminus A$.

Note that the simplification of options $A$ and $B$ (by cancellation of their common aspects) is not inconsistent with the focalization assumption (i.e., the importance of a group of arguments is that of the most important argument in the group). Rather, focalization applies once the options have been simplified so as to delete arguments that, because they are common to both options, do not contribute to making a difference between them.

The Luka case (Example 3) is a typical example of the way $\succeq^{\text{Discri}}$ outperforms $\succeq^{\text{BiPoss}}$. In that case, $A^+ = \{squash^+\}$, $A^- = \{location^{--}, price^{---}\}$, $B^+$ is empty, and $B^- = \{price^{---}\}$. Recall that $\succeq^{\text{BiPoss}}$ concludes to indifference between the two options because of the common strong cons. However, after cancelling away the $price^{---}$ argument that is present in both options, $\succeq^{\text{Discri}}$ will choose $B$, which no longer has any pros nor cons, over $A$ which is now described by a moderately strong con and a weak pro. On the Luc and Lucy cases (Examples 1 and 2), the two options do not feature any common argument, and $\succeq^{\text{Discri}}$ will therefore share the preferences of $\succeq^{\text{BiPoss}}$ (indifference in the Luc case, going to the Riviera in the Lucy case).

$\succeq^{\text{Discri}}$ is complete and quasi-transitive (its strict part, $\succ^{\text{Discri}}$ is transitive but its symmetric part is not necessarily transitive). Unsurprisingly, when $X = X^+$, $\succeq^{\text{Discri}}$ does not collapse with the max rule, but rather with the discrimax procedure (Brewka, 1989; Dubois & Fargier, 2004), that is, the comparison between $\text{OM}(A \setminus B)$ and $\text{OM}(B \setminus A)$.

As it has already been noted, $\succeq^{\text{Discri}}$ simply cancels any argument appearing in both $A$ and $B$. One could further accept the cancellation of any positive (resp. negative) argument in $A$ by another positive (resp. negative) argument in $B$, as long as these two arguments share the same order of magnitude. This yields the following $\succeq^{\text{BiLexi}}$ rule (and, later on, the $\succeq^{\text{Lexi}}$ rule), which is based on a levelwise comparison by cardinality. The arguments in $A$ and $B$ are scanned top down, until a level is reached such that the numbers of positive and negative arguments pertaining to the two alternatives are different; then, the set with the least number of negative arguments and the greatest number of positive ones is preferred. Note that it is perfectly legitimate for a qualitative decision rule to count arguments of the same strength. This simply means that one argument on one side cancels one argument of the same strength on the other side. And this is what people seem to do. What must remain qualitative is the scale expressing the *importance* of the arguments on which the decision is based; but it would be unreasonable to consider that the *number* of arguments at any given importance level is not, indeed, a number, that is, a natural integer.

Let us first define the $\lambda$-section of a set $A$ of arguments:

**Definition 8.** *For any level $\lambda \in L$:*
$A_\lambda = \{x \in A, \pi(x) = \lambda\}$ *is the $\lambda$-section of $A$.*
$A_\lambda^+ = A_\lambda \cap X^+$ *(resp. $A_\lambda^- = A_\lambda \cap X^-$ ) is its positive (resp. negative) $\lambda$-section.*

Now, a lexicographic two-sided partial ordering (called Levelwise Bivariate Tallying by Bonnefon et al.(in press)) can be introduced:

**Definition 9.**
$A \succeq^{\text{BiLexi}} B \iff |A_{\delta^*}^+| \geq |B_{\delta^*}^+|$ *and* $|A_{\delta^*}^-| \leq |B_{\delta^*}^-|$,
*where* $\delta^* = \text{Argmax}\{\lambda : |A_\lambda^+| \neq |B_\lambda^+| \text{ or } |A_\lambda^-| \neq |B_\lambda^-|\}$.

It is easy to show that $\succeq^{\text{BiLexi}}$ is reflexive, transitive, but not complete. Indeed, if at the decisive level ($\delta^*$) one of the set wins on the positive side and the other set wins on the negative side, a conflict is revealed and the procedure concludes to an incomparability.

The $\succeq^{\text{BiLexi}}$ rule is well-behaved with respect to the Lucy and Luka examples. Lucy will go to the Riviera, and Luka will chose the best located gym. In the Luka case, $\{price^{---}\}$ is preferred to $\{price^{---}, location^{--}, squash^+\}$: Once the price argument featured in both options is cancelled away, $\{location^{--}, squash^+\}$ and its moderately strong con is judged as worse than $\emptyset$. In the Luc story, the difficulty of the dilemma is clearly pointed out by the rule. Remember that option [a] involves three arguments at the highest level of importance, $landscape^{++}$, $airline^{--}$, and $price^{--}$, whilst option [b] involves only one, $governance^{--}$. Since $|\{governance^{--}\}| < |\{airline^{--}, price^{--}\}|$ while $|\{landscape^{++}\}| > |\emptyset|$, $\succeq^{\text{BiLexi}}$ concludes to incomparability, reflecting the difficulty of the decision.[5] More generally, the $\succeq^{\text{BiLexi}}$ rule concludes to incomparability if and only if there is a conflict between the positive side and the negative side at the decisive level. From a descriptive point of view, this range of incomparability is not necessarily a shortcoming of $\succeq^{\text{BiLexi}}$.

Now, if one can assume a form of compensation between positive and negative arguments of the same level within a given option, the following refinement of $\succeq^{\text{BiLexi}}$ can be obtained, that will take care of complex cases such as Luc's:

**Definition 10.**
$$A \succeq^{\text{Lexi}} B \iff \exists \lambda \in L \setminus 0_L \ such \ that \ \begin{cases} \forall \gamma > \lambda, & |A^+_\gamma| - |A^-_\gamma| & = & |B^+_\gamma| - |B^-_\gamma| \\ and & |A^+_\lambda| - |A^-_\lambda| & > & |B^+_\lambda| - |B^-_\lambda|. \end{cases}$$

Let us look one more time at Luc's dilemma. One of the strong pros of Option [a] is cancelled by one of its strong cons, and they are discarded. What remains is one strong con for each option: no option wins at this level. Therefore, the procedure then examines the next lower importance level. At this level, there are three weak pros and no con for option [b], and neither pro nor con for option [a]: Option [b] wins—and this is the first time in this article that a decision rule yields a strict preference on the Luc case. On the Lucy and Luka examples, $\succeq^{\text{Lexi}}$ breaks the ties in $\succeq^{\text{BiPoss}}$ just as $\succeq^{\text{BiLexi}}$ and $\succeq^{\text{Discri}}$ did: Lucy will go to the riviera, and Luka will chose the best located gym.

The three rules proposed in this Section obviously define monotonic bipolar set-relations. Each of them refines $\succeq^{\text{BiPoss}}$ and satisfies Preferential Independence. They can be ranked from the least to the most decisive ($\succ^{\text{Lexi}}$), which is moreover complete and transitive.

**Proposition 8.** $A \succ^{\text{BiPoss}} B \Rightarrow A \succ^{\text{Discri}} B \Rightarrow A \succ^{\text{BiLexi}} B \Rightarrow A \succ^{\text{Lexi}} B$

Both decision rules $\succ^{\text{BiLexi}}$ and $\succ^{\text{Lexi}}$ satisfy a strong form of unanimity, namely, if $A^+ \succeq B^+$ and $A^- \succeq B^-$ and moreover at least one of $A^- \succ B^-$ or $A^+ \succ B^+$ holds, then $A \succ B$ follows. However this kind of strong unanimity alone is not strong enough to achieve good and sound discrimination among options (for instance, it is the key-feature of $\succeq^{\text{Pareto}}$). In fact, $\succ^{\text{BiLexi}}$ and $\succ^{\text{Lexi}}$ satisfy much stronger efficiency properties, since $A \sim^{\text{BiLexi}} B$ only if $A$ and $B$ have the same number of positive and negative arguments at each level of importance, and $A \sim^{\text{Lexi}} B$ only if $A$ and $B$ have the same number of additional arguments of the same polarity at each level of importance after cancellation of equally important opposite arguments inside each option.

---

5. We take this opportunity to insist on the fact that incomparability should not be confused with indifference, and signals a more complex situation. In situations of incomparability, the decision-maker is perplex because not only does no alternative look better than the other, but choosing at random would not be a satisfactory way out, because any choice leads to some reason to regret. In case of indifference, both alternatives are equally attractive (or repulsive), and a random choice makes sense.

Notice that the restriction of both $\succeq^{\text{BiLexi}}$ and $\succeq^{\text{Lexi}}$ on $X^+$ amounts to the leximax preference relation (Deschamps & Gevers, 1978). We can thus use the classical encoding of the leximax (unipolar) procedure by a sum in the finite case (Moulin, 1988). A capacity is a set-function monotonic under inclusion. It is then easy to show that:

**Proposition 9.** *There exist two capacities $\sigma^+$ on $2^{X^+}$ and $\sigma^-$ on $2^{X^-}$ such that:*

$$A \succeq^{\text{Lexi}} B \Leftrightarrow \sigma^+(A^+) - \sigma^-(A^-) \geq \sigma^+(B^+) - \sigma^-(B^-)$$
$$A \succeq^{\text{BiLexi}} B \Leftrightarrow \text{ and } \begin{cases} \sigma^+(A^+) & \geq & \sigma^+(B^+) \\ \sigma^+(B^-) & \geq & \sigma^-(A^-) \end{cases}$$

For example, denoting $\lambda_1 = 0_L < \lambda_2 < \cdots < \lambda_l = 1_L$ the $l$ elements of $L$, we can use the (integer-valued) capacity:

$$\sigma^+(A^+) = \sum_{\lambda_i \in L} |A +_\lambda| \ \cdot \ |X|^i.$$

This set-function is said to be a big-stepped capacity because $|X|^i > \sum_{j<i} |X|^j$ (see Dubois & Fargier, 2004). We can similarly use a similar big-stepped capacity to represent the importance of sets of negative arguments:

$$\sigma^-(A^-) = \sum_{\lambda_i \in L} |A^-_\lambda| \ \cdot \ |X|^i.$$

This proposition means that the $\succeq^{\text{Lexi}}$ and $\succeq^{\text{BiLexi}}$ rankings are particular cases (using big-stepped probabilities) of the Cumulative Prospect Theory decision rule (Tversky & Kahneman, 1992). See Section 6 for further discussion.

In summary, $\succeq^{\text{Lexi}}$ complies with the spirit of qualitative bipolar reasoning while being efficient. In the meantime, it has the advantages of numerical measures (transitivity and representability by a pair of functions). Finally, in order to make a full case for the attractive $\succeq^{\text{Lexi}}$, we will show that $\succeq^{\text{Lexi}}$ is the only rule that at the same time (i) refines $\succeq^{\text{BiPoss}}$, (ii) is a weak order, and (iii) satisfies the principle of preferential independence without introducing any bias in the order on $\mathbb{X}$.

**Definition 11.** *A refinement $\succeq'$ of a monotonic bipolar set relation $\succeq$ is unbiased if it preserves the ground relation of the latter: $\succeq'_{\mathbb{X}} \ \equiv \ \succeq_{\mathbb{X}}$.*

In order to prove our claim, let us first establish the following proposition that applies to any transitive relation satisfying preferential independence (note that completeness is not required). This proposition establishes a principle of anonymity, according to which two equivalent subsets are interchangeable provided that they do not overlap the other sets of arguments:

**Proposition 10.** *Let $\succeq$ be a transitive monotonic bipolar set-relation that satisfies preferential independence. Then for any $A, B, C, D$ such that $A \cap (C \cup D) = \emptyset$ and $C \sim D$:*

- $A \cup C \succeq B \iff A \cup D \succeq B$;

- $B \succeq A \cup C \iff B \succeq A \cup D$.

A direct consequence of this property it that a set of arguments that cancel each other can be added to another set without changing the preferences involving the latter (just let $D \sim \emptyset$):

**Corollary 1.** *Let $\succeq$ be a transitive monotonic bipolar set-relation that satisfies preferential independence. Then for any $A, B, C$ such that $A \cap C = \emptyset$, $C \sim \emptyset$,*

- $A \succeq B \iff A \cup C \succeq B$;

- $B \succeq A \iff B \succeq A \cup C$.

Another noteworthy consequence of Proposition 10 is an extended principle of preferential independence:

**Corollary 2.** *Let $\succeq$ be a transitive monotonic bipolar set-relation that satisfies preferential independence. Then for any $A, B, C, D$ such that $(A \cup B) \cap (C \cup D) = \emptyset$ and $C \sim D$:*

$$A \succeq B \iff A \cup C \succeq B \cup D$$

Finally, Proposition 10 allows to show that[6]:

**Corollary 3.** *Let $\succeq$ be a transitive monotonic bipolar set-relation that satisfies preferential independence. Then for any $A, B \subseteq X$, $x, y \in X$ such that $x \notin A, y \notin B$, $\{x\} \sim \{y\}$:*

$$A \succeq B \iff A \cup \{x\} \succeq B \cup \{y\}$$

The three above results are instrumental when proving the third representation result of the paper:

**Theorem 2.** *Let $\succeq$ be a monotonic bipolar set-relation and $\pi$ be an order-of-magnitude distribution over the elements of $X$. The following propositions are equivalent:*

1. *$\succeq$ is complete, transitive, satisfies preferential independence and is an unbiased refinement of $\succeq^{\mathrm{BiPoss}}$;*

2. *$\succeq \ \equiv \ \succeq^{\mathrm{Lexi}}$.*

This theorem concludes our argumentation in favor of $\succeq^{\mathrm{Lexi}}$, which is shown to be the only rule to satisfy the natural principles of order-of-magnitude reasoning while being decisive and practical (e.g., completeness, transitivity, representability by a pair of functions).

## 6. Related Works

In this section, we relate our decision rules and our general approach to several traditions of research, in Cognitive Psychology as well as in Artificial Intelligence.

An interesting connection can be made between our decision rules and the *Take the Best* heuristic that has been extensively studied by psychologists since its introduction by

---

6. Note that, in Corollary 3, neither condition $\{x\} \cap B = \emptyset$ nor condition $\{y\} \cap A = \emptyset$ are required; this is the main difference with Corollary 2; on the other hand, Corollary 3 is restricted to the addition of singletons.

Gigerenzer and Goldstein (1996). Take the Best is a so-called "fast and frugal heuristic" for comparing two objects based on their values in a series of binary cues. It is fast and frugal because it focuses on a limited subset of the available information in order to make a decision; and it has heuristic value because it shows reasonable accuracy as compared to less frugal comparison rules (see Gigerenzer et al., 1999; Katsikopoulos & Martignon, 2006, for simulations and empirical tests). As a consequence, evolutionary psychologists have argued that Take the Best is an adapted strategy for the kind of binary cue-based comparison we are investigating.

Take the Best requires that all arguments under consideration have different orders of magnitude, so that they can be ranked lexicographically: there does not exist $x$, $y$ such that $\pi(x) = \pi(y)$. Furthermore, each argument is considered to generate its polar opposite: if the pro $x^+$ is attached to an option [a], then the con $x^-$ such that $\pi(x^-) = \pi(x^+)$ is automatically attached to all options that do not feature $x^+$ (and reciprocally). Then applying Take the Best amounts to scanning the arguments top-down, starting from the most important, and to stop as soon as an argument is found that favors one option but not the other. Interestingly, when applied to such linearly ranked arguments, the $\succeq^{\text{Discri}}$, $\succeq^{\text{BiLexi}}$, and $\succeq^{\text{Lexi}}$ rules coincide with the Take the Best heuristic. But the new decision rules proposed here are able to account for more choice situations than the Take the Best heuristic—e.g., several criteria can share the same degree of importance. In this sense, they are natural extensions of the Take the Best qualitative rule advocated by psychologists.

Originally a descriptive model of risky decisions, Cumulative Prospect Theory (Tversky & Kahneman, 1992) provides another psychological account of decisions involving positive and negative arguments. Contrary to the Take the Best approach, it is oriented towards the quantitative evaluation of decisions. The key notions of Cumulative Prospect Theory are that individuals assess outcomes relatively to some reference point, rather than absolutely; that they are more concerned with losses than with gains, ceteris paribus; and that individuals overweight extreme outcomes that occur with a small probability.

More technically, Cumulative Prospect Theory assumes that the reasons supporting a decision and the reasons against it can be measured by means of two capacities $\sigma^+$ and $\sigma^-$; $\sigma^+$ reflects the importance of the group of positive arguments, $\sigma^-$ the importance of the group of negative arguments. The higher $\sigma^+$, the more convincing the set of positive arguments; and conversely, the higher $\sigma^-$, the more deterring the set of negative arguments. Furthermore, Cumulative Prospect Theory assumes that it is possible to map these evaluations to a so-called "net predisposition" score, expressed on a single scale:

$$\forall A \subseteq X, \text{NP}(A) = \sigma^+(A^+) - \sigma^-(A^-),$$

where $A^+ = A \cap X^+, A^- = A \cap X^-$. Alternatives are then ranked according to this net predisposition: $A \succeq^{\text{NP}} B \iff \sigma^+(A^+) - \sigma^-(A^-) \geq \sigma^+(B^+) - \sigma^-(B^-)$. Proposition 9 (Section 5) actually shows that $\succeq^{\text{Lexi}}$ is a particular case of this rule (using big-stepped capacities). Interestingly, the $\succeq^{\text{BiPoss}}$ rule is not foreign to the comparison of net predispositions. More precisely, it can be viewed as the ordinal counterpart to net predisposition comparison. Let us first write $A \succeq^{\text{NP}} B$ as $\sigma^+(A^+) + \sigma^-(B^-) \geq \sigma^+(B^+) + \sigma^-(A^-)$. Now, it is immediate that changing + into max, and changing $\sigma^+$ and $\sigma^-$ in this inequality into possibility measures, yields the $\succeq^{\text{BiPoss}}$ rule.

Cumulative Prospect Theory and its variants assume some kind of independence between $X^+$ and $X^-$, but this assumption does not always hold. *Bi-capacities* (Grabisch & Labreuche, 2002, 2005) were introduced to handle such non-separable bipolar preferences: a measure $\sigma$ is defined on $Q(X) := \{(U, V) \in 2^X, U \cap V = \emptyset\}$, and increases (resp., decreases) with the addition of elements in $U$ (resp., in $V$)[7]. Bi-capacities originally stemmed from bi-cooperative games (Bilbao, Fernandez, Jiménez Losada, & Lebrón, 2000), where players are divided into two groups, the "pros" and the "cons": a player $x$ is sometimes a pro, sometimes a con, but cannot be both simultaneously. In the context of bipolar decision, we typically set $U = A^+$ and $V = A^-$ and measure the attractivity of $A$ by $\sigma(A^+, A^-)$. The net predisposition of Cumulative Prospect Theory is recovered by letting $\sigma(A^+, A^-) = \sigma^+(A^+) - \sigma^-(A^-) = \mathrm{NP(A)}$.

Since the comparison of net predispositions (and, more generally, bi-capacities or bi-cooperative games) systematically provides a complete and transitive preference, it can fail to capture a large range of decision-making attitudes: contrasting affects make decision difficult, so why should the comparison of objects having bipolar evaluations systematically yield a complete relation? It might imply some incompatibilities. That is why bi-capacities were generalized by means of bipolar capacities (Greco et al., 2002). The idea underlying bipolar capacities is to use two measures, a measure of positiveness (that increases with the addition of positive arguments and the deletion of negative arguments) and a measure of negativeness (that increases with the addition of negative arguments and the deletion of positive arguments), *but not to combine them.* In other terms, a bipolar capacity $\sigma$ can be equivalently defined by a pair of bi-capacites $\sigma^+$ and $\sigma^-$, namely by: $\sigma(A) = (\sigma^+(A^+, A^-), \sigma^-(A^-, A^+))$. Then $A$ is preferred to $B$ with respect to $\sigma$ if and only if $A$ is preferred to $B$ with respect to both $\sigma^+$ and $\sigma^-$— that is, according to the sole Pareto principle. This allows for the representation of conflicting evaluations and leads to a partial order.

Our approach provides clear qualitative counterparts of Cumulative Prospect Theory, bi-capacities, and bipolar capacities only to a certain extent. Indeed, $\succeq^{\mathrm{Lexi}}$ belongs to the Cumulative Prospect Theory family, and can thus be represented by a bi-capacity, or more generally a bipolar capacity. In contrast, $\succeq^{\mathrm{Pareto}}$ obviously belongs to the family of bipolar capacities. But rules $\succeq^{\mathrm{BiPoss}}$, $\succeq^{\mathrm{Discri}}$, $\succeq^{\mathrm{BiLexi}}$ and $\succeq^{\mathrm{Impl}}$ are not in the spirit of bi-capacities nor bi-cooperative games. First of all, and contrary to these models, some of these decision rules do not provide a complete preorder between alternatives – they argue on the contrary that conflicts between positive and negative arguments of the same strength should lead to a conflict, thus to incomparable alternatives. Other decision rules, like $\succeq^{\mathrm{BiPoss}}$ cannot be understood as a comparison of some $\sigma(A^+, A^-)$ to some $\sigma(B^+, B^-)$; this rule indeed compares $\max(\mathrm{OM}(A^+), \mathrm{OM}(B^-))$ to $\max(\mathrm{OM}(B^+), \mathrm{OM}(A^-))$. For $\succeq^{\mathrm{BiPoss}}$, the right scheme would rather be the comparison of some $\sigma(A^+, B^-)$ to some $\sigma(B^+, A^-)$. This suggests that neither the framework of bi-cooperative games, nor the ones of bi-capacities nor of bipolar capacities are yet general enough.

---

7. For a bibliography about bi-capacities and bi-cooperative games see the work of Grabisch and colleagues (Grabisch & Labreuche, 2005; Grabisch & Lange, 2007). Most of the developments about these notions concern the computation of their Shapley value, or the definition of Choquet integrals. But apart from Cumulative Prospect Theory, very few instances of this general framework are presented. The point is probably that, because it provides a complete and transitive comparison, it cannot highlight the presence of conflicting information.

Finally, notice that we have couched our results in a terminology borrowing to argumentation and decision theories, and indeed we consider they can be relevant for both. The paper is also relevant to argumentative *reasoning* for the evaluation of sets of arguments in inference processes (Cayrol & Lagasquie-Schiex, 2005). The connection between reasoning and decision in the setting of argumentation is laid bare by Amgoud, Bonnefon, and Prade (2005). They propose an extensive framework where arguments are constructed from a knowledge base and a base of logically defined criteria. Arguments are evaluated in terms of certainty, strength and degree of attainment of the corresponding criterion. Assuming a common scale for these three aspects, they separately compare individual positive arguments, and negative ones, using a simple aggregation of these weights, but never compare arguments of different polarity. The idea of using logical arguments to compare decisions is first discussed in the setting of possibilistic logic by Amgoud and Prade (2004). Amgoud and Prade (2006) further elaborated this framework, where the explicit use of a knowledge base and a goal base in the construction of arguments builds a bridge between argumentative reasoning and qualitative decision under uncertainty.

## 7. Conclusion

This paper has focused on a particular class of bipolar decision making situations, namely those that are qualitative rather than quantitative. What we have proposed is an extension of possibility theory to the handling of sets containing arguments considered as positive or negative. We have laid bare the importance of $\succeq^{\text{Lexi}}$ and $\succeq^{\text{BiLexi}}$ as decision rules for qualitative bipolar decision-making. Both rules separately evaluate the positive and negative sets of arguments, by means of big-stepped capacities $\sigma^+$ and $\sigma^-$. Then, the $\succeq^{\text{Lexi}}$ rule aggregates the two measures in agreement with Cumulative Prospect Theory's "net predisposition." In contrast, the $\succeq^{\text{BiLexi}}$ rule does not merge the positive and negative measures, allowing for the expression of conflict and incomparability. In a sense, these two rules combine the best of two worlds: while they agree with the spirit of order of magnitude reasoning, they are more decisive and efficient than basic rules such as $\succeq^{\text{BiPoss}}$, and they offer the same practical advantages as the quantitative Cumulative Prospect Theory—e.g., transitivity and representability by a pair of functions.

This paper has adopted a *prescriptive* point of view in the sense that the rules were studied with respect to the properties that a qualitative theory of bipolar decision-making should obey. The representation theorems of Sections 4 and 5 show that the use of $\succeq^{BiPoss}$ and $\succeq^{\text{Lexi}}$ are the only well-behaved ones when dealing with qualitative bipolar information. In parallel, we tested the *descriptive* power of our rules, i.e., their accuracy when predicting the behavior of human decision makers. Our experimental results (Bonnefon et al., in press; Bonnefon & Fargier, 2006) confirm $\succ^{\text{BiPoss}}$ as a basic ordinal decision-making rule: when $\succ^{\text{BiPoss}}$ yields a strict preference, it is very uncommon that human decision-makers disagree with this preference. Furthermore, results strongly suggest that $\succeq^{\text{Lexi}}$ is the decision rule generally followed by decision-makers: $\succeq^{\text{Lexi}}$ accurately predicted nearly 80% of the 2,000 decisions we collected, and was always the one decision-makers individually agreed with the most. Finally, results suggest that human decision-makers do sometimes find some decisions incomparable; and when they do, it is usually in situations when $\succeq^{\mathrm{BiLexi}}$ would predict so.

In the present paper, we did not address the question of the computational complexity of comparing two objects with a bipolar decision rule. Actually, the presence of bipolar information does not change the range of complexity of the comparison of objects. A detailed complexity study is out of the scope of the paper, but we can at least say that for all the rules presented in this paper, the comparison of two alternatives is at most polynomial in the number of criteria in $X$. This comparison indeed relies on the computation of the strength of four subsets of $X$ – computation that is sometimes preceded by a simple deletion step (for the rules based on a discri- or a lexi- comparison). The computation of the strength of a set is itself linear, since performed by means of an aggregation operator. The complexity of comparing two alternatives with the $\succeq^{\mathrm{BiPoss}}$ or the $\succeq^{\mathrm{Lexi}}$ rules is for instance in $O(Card(X))$ (it can be computed as a simple aggregation of individual strengths). Finally, notice that all our rules provide a transitive strict preference, so no cycle can appear as it may be the case in CP-nets. If combinatorial alternatives were to be considered, we could easily use a branch and bound algorithm for looking for the best alternative(s); the problem of optimization with a bipolar aggregation is not harder than with a unipolar one: the corresponding decision problem remains NP-complete (Fargier & Wilson, 2007).

This concludes our study of qualitative bipolar reasoning. However, we are aware of some unresolved issues this paper has raised. For example, Section 6 suggests that the framework of bi-capacities, and even bipolar capacities, is not rich enough to accommodate the full range of qualitative bipolar rules we have introduced. Secondly, our results were established within a restricted framework, where a relevant criterion is either a complete pro or a complete opponent w.r.t. each decision, in the spirit of bi-cooperative games. This is clearly a simpler approach than usual multi-criteria decision-making frameworks, where each $x \in X$ is a full-fledged criterion rated on a bipolar utility scale like $L_x = [-1_x, +1_x]$, containing a neutral value $0_x$. Thus, a natural extension of the present work would be to address qualitative bipolar criteria whose satisfaction is a matter of degree.

## Appendix A. Proofs

To simplify notations, let $a^+ = \mathrm{OM}(A^+)$, $a^- = \mathrm{OM}(A^-)$, $b^+ = \mathrm{OM}(B^+)$, $b^- = \mathrm{OM}(B^-)$, $c^+ = \mathrm{OM}(C^+)$, $c^- = \mathrm{OM}(C^-)$. Hence $\mathrm{OM}(A) = \max(a^+, a^-)$, $\mathrm{OM}(A^+ \cup B^-) = \max(a^+, b^-)$ and so on.

*Proposition 1.* The proof of completeness is trivial, as $\mathrm{OM}(A^+ \cup B^-)$ and $\mathrm{OM}(B^+ \cup A^-)$ can always be compared. To prove the transitivity of $\succ^{\mathrm{BiPoss}}$, let us assume $\max(a^+, b^-) > \max(a^-, b^+)$ and $\max(b^+, c^-) > \max(b^-, a^+)$. Then, letting $b = \max(b^+, b^-)$, we get $\max(a^+, b, c^-) > \max(a^-, b, c^+)$. As a consequence, $\max(a^+, c^-) > b$. Hence, $\max(a^+, c^-) > \max(a^-, b, c^+) \geq \max(a^-, c^+)$. $\qquad\square$

*Proposition 2.* Quasi-transitivity is proved in Proposition 1. Positive and negative monotony, as well as SQC, follow from the monotony of OM, which is a possibility measure: i.e., $\forall U, V, \mathrm{OM}(U \cup V) \geq \mathrm{OM}(V)$. Non-triviality of $\succeq^{\mathrm{BiPoss}}$ is obtained from the non-triviality of $\pi$ (there exists $x$ such as $\pi(x) = 0$), which implies $\mathrm{OM}((X^+)^+ \cup (X^-)^-) = \mathrm{OM}(X^+ \cup$ $X^-) > 0$ while $\mathrm{OM}((X^+)^- \cup (X^-)^+) = \mathrm{OM}(\emptyset) = 0$. Clarity of argument is also trivial: if $x \in X^+$ then $\mathrm{OM}(\{x\}^+ \cup (\emptyset)^-) = \pi(x) \geq \mathrm{OM}(\{x\}^- \cup (\emptyset)^+) = \mathrm{OM}(\emptyset) = 0$: $\{x\} \succeq^{\mathrm{BiPoss}} \emptyset$. If $x$ is a con, we get in the same way $\{x\} \preceq^{\mathrm{BiPoss}} \emptyset$. If $x$ has null importance, we get $\mathrm{OM}(\{x\}^+ \cup (\emptyset)^-) = \mathrm{OM}(\{x\}^- \cup (\emptyset)^+) = \mathrm{OM}(\emptyset) = 0$: $\{x\} \sim^{\mathrm{BiPoss}} \emptyset$. To prove weak unanimity, suppose that $A^+ \succeq^{\mathrm{BiPoss}} B^+$ and $A^- \succeq^{\mathrm{BiPoss}} B^-$. Obviously $A^+ \succeq^{\mathrm{BiPoss}} B^+ \iff \mathrm{OM}(A^+) \geq \mathrm{OM}(B^+)$ and $A^- \succeq^{\mathrm{BiPoss}} B^- \iff \mathrm{OM}(B^-) \geq \mathrm{OM}(A^-)$. Hence $\mathrm{OM}(A^+ \cup B^-) \geq \mathrm{OM}(B^+ \cup A^-) : A \succeq^{\mathrm{BiPoss}} B$. $\qquad \square$

*Proposition 3.* When restricted to singletons, $\succeq^{\mathrm{BiPoss}}$ is a weak order that ranks the positive arguments by decreasing values of $\pi$, then the null arguments ($\pi = 0$), then the negative arguments by increasing value of $\pi$. This ranking defines a complete and transitive relation. This proves that the relation $\succeq_\mathbb{X}$ induced by $\succeq^{\mathrm{BiPoss}}$ is a weak order. Axioms POSC is easy to check, since $\{x^+, y^-\} \sim \emptyset$ (resp, $\{z^+, y^-\} \sim^{\mathrm{BiPoss}} \emptyset$) if and only if $\pi(x^+) = \pi(y^-)$ (resp. $\pi(z^+) = \pi(y^-)$). Then $\{x^+, y^-\} \sim^{\mathrm{BiPoss}} \emptyset$ and $\{z^+, y^-\} \sim \emptyset$ imply $\pi(x^+) = \pi(z^+)$. So, $\{x^+\} \sim^{\mathrm{BiPoss}} \{z^+\}$, i.e., $x^+ \sim_\mathbb{X} z^+$. The proof of NEGC is similar. The $\mathbb{X}$-monotony of $\succeq^{\mathrm{BiPoss}}$ is shown as follows. Let $A, x, x'$ be such that $A \cap \{x, x'\} = \emptyset$ and $x' \succeq_\mathbb{X} x$. Three cases are possible:

1. $x \in X^+$. Then $x' \in X^+$ and $\pi(x') \geq \pi(x)$.

   - if $A \cup \{x\} \succ^{\mathrm{BiPoss}} B$: then $\mathrm{OM}(A^+ \cup \{x\} \cup B^-) > \mathrm{OM}(A^- \cup B^+)$. Since $\pi(x') \geq \pi(x)$, we get $\mathrm{OM}(A^+ \cup \{x'\} \cup B^-) \mathrm{OM}(A^- \cup B^+)$: $A \cup \{x'\} \succ^{\mathrm{BiPoss}} B$.

   - if $A \cup \{x\} \sim^{\mathrm{BiPoss}} B$: then $\mathrm{OM}(A^+ \cup \{x\} \cup B^-) = \mathrm{OM}(A^- \cup B^+)$. Replacing $x$ by $x'$, i.e., $\pi(x)$ by $\pi(x')$, the first OM level increases, so we get $\mathrm{OM}(A^+ \cup \{x'\} \cup B^-) \geq \mathrm{OM}(A^- \cup B^+)$: $A \cup \{x'\} \succeq^{\mathrm{BiPoss}} B$.

   - if $B \succ A \cup \{x'\}$, then $\mathrm{OM}(B^+ \cup A^-) > \mathrm{OM}(A^+ \cup \{x'\} \cup B^-)$. Replacing $x'$ by $x$, i.e., $\pi(x')$ by $\pi(x)$, the second OM level decreases, so we get $\mathrm{OM}(B^+ \cup A^-) > \mathrm{OM}(A^+ \cup \{x\} \cup B^-)$, i.e. $B \succ^{\mathrm{BiPoss}} A \cup \{x\}$.

   - if $B \sim A \cup \{x'\}$, then $\mathrm{OM}(B^+ \cup A^-) = \mathrm{OM}(A^+ \cup \{x'\} \cup B^-)$. Replacing $x'$ by $x$, i.e. $\pi(x')$ by $\pi(x)$, the second OM level decreases, so $\mathrm{OM}(B^+ \cup A^-) \geq \mathrm{OM}(A^+ \cup \{x\} \cup B^-)$, i.e. $B \succeq^{\mathrm{BiPoss}} A \cup \{x\}$.

2. $x' \in X^-$. Then $x \in X^-$ and $\pi(x) \geq \pi(x')$. The same kind of four-case proof can be carried out.

3. $x' \in X^+$ **and** $x \in X^-$.

   - if $A \cup \{x\} \succ^{\mathrm{BiPoss}} B$, i.e., $\mathrm{OM}(A^+ \cup B^-) > \mathrm{OM}(A^- \cup B^+ \cup \{x\})$, it follows that $\mathrm{OM}(A^+ \cup B^- \cup \{x'\}) > \mathrm{OM}(A^- \cup B^+)$, so $A \cup \{x'\} \succ^{\mathrm{BiPoss}} B$.

   - Similarly, if $A \cup \{x\} \sim^{\mathrm{BiPoss}} B$, i.e., $\mathrm{OM}(A^+ \cup B^-) = \mathrm{OM}(A^- \cup B^+ \cup \{x\})$, it follows that $\mathrm{OM}(A^+ \cup B^- \cup \{x'\}) \geq \mathrm{OM}(A^- \cup B^+ \cup \{x\})$, and then $\mathrm{OM}(A^+ \cup B^- \cup \{x'\}) \geq \mathrm{OM}(A^- \cup B^+ \cup \{x\})$, which implies $\mathrm{OM}(A^+ \cup B^- \cup \{x'\}) \geq \mathrm{OM}(A^- \cup B^+)$. Hence ($x'$ being positive) $A \cup \{x'\} \succeq^{\mathrm{BiPoss}} B$.

   - if $B \succ A \cup \{x'\}$, which means $\mathrm{OM}(B^+ \cup A^-) > \mathrm{OM}(A^+ \cup \{x'\} \cup B^-)$ since $x'$ is positive. Obviously, $\mathrm{OM}(B^+ \cup A^- \cup \{x\}) \geq \mathrm{OM}(B^+ \cup A^-)$ and $\mathrm{OM}(A^+ \cup \{x'\} \cup B^-) \geq \mathrm{OM}(A^+ \cup B^-)$. Hence $\mathrm{OM}(B^+ \cup A^- \cup \{x\}) > \mathrm{OM}(A^+ \cup B^-)$: $B \succ^{\mathrm{BiPoss}} A \cup \{x\}$.

- Similarly, if $B \sim^{\text{BiPoss}} A \cup \{x'\}$ we have: $\text{OM}(B^+ \cup A^- \cup \{x\}) \geq \text{OM}(B^+ \cup A^-) = \text{OM}(A^+ \cup \{x'\} \cup B^-) \geq \text{OM}(A^+ \cup B^-)$: $B \succeq^{\text{BiPoss}} A \cup \{x\}$.

$\square$

*Proposition 4.*

$\succeq^{\text{BiPoss}}$ **satisfies GCLO** Recall that $A \succeq^{\text{BiPoss}} B \iff \max(a^+, b^-) \geq \max(b^+, a^-)$ and $C \succeq^{\text{BiPoss}} D \iff \max(c^+, d^-) \geq \max(d^+, c^-)$. Hence, $\max(a^+, b^-, c^+, d^-) \geq \max(b^+, a^-, d^+, c^-)$, i.e., $A \cup C \succeq^{\text{BiPoss}} B \cup D$.

$\succ^{\text{BiPoss}}$ **satisfies GNEG** Recall that $A \succ^{\text{BiPoss}} B \iff \max(a^+, b^-) > \max(b^+, a^-)$ and $C \succ^{\text{BiPoss}} D \iff \max(c^+, d^-) > \max(d^+, c^-)$. Hence: $\max(a^+, b^-, c^+, d^-) > \max(b^+, a^-, d^+, c^-)$, i.e., $A \cup C \succ^{\text{BiPoss}} B \cup D$.

$\square$

*Theorem 1.* Let us first build $\pi$. By CA, any singleton $\{x\}$ is comparable to $\emptyset$. So $X^+ = \{x, \{x\} \succ \emptyset\}$, $X^- = \{x, \{x\} \prec \emptyset\}$ and $X^0 = \{x, \{x\} \sim \emptyset\}$ are soundly defined.

Let us define a relation $\geq$ on $X$ as follows:

- $x, y \in X^0 : x \geq y$ and $y \geq x$.

- $x, y \in X^+ : x \geq y \iff \{x\} \succeq \{y\}$

- $x, y \in X^- : x \geq y \iff \{y\} \succeq \{x\}$

- $x \in X^+, y \in X^- : x \geq y \iff not(\emptyset \succ \{x, y\})$

- $x \in X^+, y \in X^0 : x > y$

- $x \in X^-, y \in X^0 : x > y$

Because of axiom CA, $X^+$, $X^-$ and $X^0$ are disjoint. The previous definition is thus well founded. $\geq$ is complete by definition. We prove now that $\geq$ is transitive. Suppose that $x \geq y$ and $y \geq z$ and let us perform the following case analysis:

- $x, y, z \in X^+$: Then $x \geq z$ is trivial because $\succeq_{\mathbb{X}}$ is transitive and can be identified with $\geq$ within $X^+$.

- $x, y, z \in X^-$: Then $x \geq z$ is trivial because $\succeq_{\mathbb{X}}$ is transitive and can be identified with $\geq$ within $X^-$.

- $x \in X^0$: then $x \geq y$ means by STQ that $y$ is also in $X^0$, which in turn implies $z \in X^0$. So, by STQ again, $x \equiv z$.

- $y \in X^0$: $y \geq z$ implies by STQ that $z$ is also in $X^0$. So, by STQ again, $x \geq z$.

- $z \in X^0$: $x \geq z$ is always true (by status quo consistency again).

- $x \in X^+$, $y \in X^-$ and $z \in X^-$. Then by definition, $x \geq y$ means $\{x, y\} \succeq \emptyset$ and $y \geq z$ means $\{z\} \succeq \{y\}$. By $\mathbb{X}$-monotony we can replace $y$ by $z$ without reversing the preference: $\{x, z\} \succeq \emptyset$, i.e. $x \geq z$.

- $x \in X^+$, $y \in X^+$ and $z \in X^-$ : the proof of $x \geq z$ is similar to the one in the previous item.

- $y \in X^+$ $x \in X^-$ and $z \in X^-$. Suppose that $\{x, y\} \preceq \emptyset$ $(x \geq y)$, $\{z, y\} \succeq \emptyset$ $(y \geq z)$ and $\{x\} > \{z\}$ $(z > x$ for negative arguments). If $\{z, y\} \succ \emptyset$, then $\mathbb{X}$-monotony implies $\{x, y\} \succ \emptyset$ (thus a contradiction). If $\emptyset \succ \{x, y\}$ then $\mathbb{X}$-monotony implies $\emptyset \succ \{z, y\}$ (second contradiction). Last case, if $\{x, y\} \sim \emptyset$ and $\{z, y\} \sim \emptyset$, then POSC implies $\{x\} \sim \{z\}$ (last contradiction). So, $z > x$ does not hold and thus, by completeness of $\geq$, $x \geq z$.

- $y \in X^-$ $x \in X^+$ and $z \in X^+$. The proof of $x \geq z$ is similar to the one in the previous item, using NEGC instead of POSC.

So, $\geq$ is a weak order. It can be encoded by a distribution $\pi : X \mapsto [0_L, 1_L]$, $[0_L, 1_L]$ being a totally ordered scale. Level $0_L$ is mapped to elements of $X_0$. Now, we have to show the equivalence between $\succeq$ and the relation $\succeq^{\text{BiPoss}}$ induced from $\pi$, namely to prove that $A \succeq B \iff \text{OM}(A^+ \cup B^-) \geq \text{OM}(A^- \cup B^+)$. Since $\succeq$ is complete, this amounts to showing that $\text{OM}(A^+ \cup B^-) = \text{OM}(A^- \cup B^+)$ implies $A \sim B$ and that $\text{OM}(A^+ \cup B^-)\text{OM}(A^- \cup B^+)$ implies $A \succ B$. Let us first prove that $\text{OM}(A^+ \cup B^-)\text{OM}(A^- \cup B^+)$ implies $A \succ B$. Suppose that the element of highest $\pi$ value in $A^+ \cup B^-$ is $x \in A^+$. So, for any $y \in A^-$, $\{x, y\} \succ \emptyset$ and $w \in B^+$, $\{x\} \succ \{w\}$. By GNEG we get $\{x\} \cup A^- \succ B^+$ and positive and negative monotony then imply $A^+ \cup A^- \succ B^+ \cup B^-$. If the element of highest $\pi$ value in $A^+ \cup B^-$ were some $v \in B^-$ then for any $w \in B^+$, $\{w, v\} \prec \emptyset$ and for any $y \in A^-$, $\{y\} \succ \{v\}$. By GNEG we get $\{v\} \cup B^+ \prec A^-$ and positive and negative monotony then imply $A^+ \cup A^- \succ B^+ \cup B^-$. Let us now prove that $\text{OM}(A^+ \cup B^-) \geq \text{OM}(A^- \cup B^+)$ implies $A \succeq B$. Suppose that the element of highest $\pi$ value in $A^+ \cup B^-$ is $x \in A^+$. So, for any $y \in A^-$, $\{x, y\} \succeq \emptyset$ and $w \in B^+$, $\{x\} \succeq \{w\}$. By GCLO we get $\{x\} \cup A^- \succeq B^+$ and positive and negative monotony then imply $A^+ \cup A^- \succeq B^+ \cup B^-$. If the element of highest $\pi$ value in $A^+ \cup B^-$ were some $v \in B^-$ then for any $w \in B^+$, $\{w, v\} \preceq \emptyset$ and for any $y \in A^-$, $\{y\} \succeq \{v\}$. By GCLO we get $\{v\} \cup B^+ \preceq A^-$ and positive and negative monotony then implies $A^+ \cup A^- \succeq B^+ \cup B^-$. $\qquad \square$

*Proposition 5.* Consider the four cases identified in the text. Clearly situation 1 corresponds to a case where $A \sim^{\text{Impl}} B$. In case 4, strict dominance $A \succ^{\text{Impl}} B$ prevails. Case 2 may lead to three different conclusions. Equivalence arises when $a^+ = b^+ \max(a^-, b^-)$: the cons being of low level w.r.t to the pros, they are not taken into account and indifference prevails based on the pros. If $a^+ = b^+ = b^- > a^-$, then $A \succeq^{\text{Impl}} B$ holds but not $B \succeq^{\text{Impl}} A$. So, $A \succ^{\text{Impl}} B$ since the arguments against $A$ are too weak. By symmetry, $a^+ = b^+ = a^- > b^+$ concludes to $B \succ^{\text{Impl}} A$. Case 3 is handled in a similar manner: equivalence arises when $a^- = b^- > \max(a^+, b^+)$; if $a^- = b^- = a^+ > b^+$, $A \succ^{\text{Impl}} B$ holds since the arguments for $B$ are too weak; if $a^- = b^- = b^+ > a^+$, $B \succ^{\text{Impl}} A$ Finally, $A \bowtie^{\text{Impl}} B$ is concluded when neither $A \succeq^{\text{Impl}} B$ nor $B \succeq^{\text{Impl}} A$. This arises in two cases only, when $a^+ = a^- > \max(b^-, b^+)$ or in the symmetric case $b^+ = b^- \max(a^-, a^+)$. $\qquad \square$

*Proposition 6.* Assume $A \succeq^{\text{Impl}} B$ and $B \succeq^{\text{Impl}} C$. Using the above conventions, consider:

1. Any of the following situations ensuring $A \succeq^{\text{Impl}} B$:

    (a) $a^+ = b^+ = a^- = b^-$;

    (b) $a^+ = b^+ \geq \max(a^-, b^-)$;

    (c) $a^- = b^- \geq \max(a^+, b^+)$;

    (d) $\max(a^+, b^-) > \max(a^-, b^+)$.

2. And any of the following situations ensuring $B \succeq^{\text{Impl}} C$ the following group:

    (a) $b^+ = c^+ = b^- = c^-$;

    (b) $b^+ = c^+ \geq \max(b^-, c^-)$;

    (c) $b^- = c^- \geq \max(b^+, c^+)$;

    (d) $\max(b^+, c^-) > \max(b^-, c^+)$.

Combining one condition in the first group with one in the second group yields one of the corresponding conditions ensuring $A \succeq^{\text{Impl}} C$. For instance,

- Combining conditions 1b and 2b yields $a^+ = c^+ \geq \max(a^-, b^-, c^-) \geq \max(a^-, c^-)$.

- Combining conditions 1b and 2c yields $a^+ \geq \max(a^-, c^-)$ and $c^- \geq \max(a^+, c^+)$. Hence $a^+ \geq \max(a^-, a^+, c^+)$, $c^- \geq \max(a^-, c^-, c^+)$ and $a^+ = c^-$. Hence $a^+ = c^- \geq \max(a^-, c^+)$. If the inequality is strict this is condition d. If $a^+ = c^- = \max(a^-, c^+)$, this is condition b or c.

- Combining condition 1d and condition 2b yields $\max(a^+, b^-) > \max(a^-, c^+)$ and $c^+ \geq \max(b^-, c^-)$. Hence $a^+ > \max(a^-, c^+)$.

The other cases can be handled similarly. $\qquad\blacksquare$

*Proposition 7.* $A \succ^{\text{BiPoss}} B$ if and only if $\max(a^+, b^-) > \max(a^-, b^+)$ which thanks to Proposition 5 implies $A \succ^{\text{Impl}+} B$. $\qquad\blacksquare$

*Proposition 8.*

- Suppose that $A \succ^{\text{BiPoss}} B$: then $\exists x^* \in A^+ \cup B^-$ such that $\forall x \in A^- \cup B^+, \pi(x^*)\pi(x)$. If many arguments satisfy this property, let $x^*$ be one of those maximizing $\pi$. $x^*$ is thus not in $A^- \cup B^+$. So, it is either in $A^+ \setminus B^+$ or in $B^- \setminus A^-$. Since $A^+ \setminus B^+ = (A \setminus B)^+$ and $B^- \setminus A^- = (B \setminus A)^-$, we can write $x^* \in (A \setminus B)^+ \cup (B \setminus A)^-$. On the other hand $(B \setminus A)^+ \cup (A \setminus B)^- \subseteq A^- \cup B^+$ and no element in $A^- \cup B^+$ has a higher degree than $x^*$. So, $\text{OM}((B \setminus A)^+ \cup (A \setminus B)^-) < \pi(x^*)$. So, $A \succ^{\text{Discri}} B$. This proves that $\succeq^{\text{Discri}}$ refines $\succeq^{\text{BiPoss}}$.

- Suppose that $A \succ^{\text{Discri}} B$: then $\exists x^* \in A^+ \setminus B^+ \cup B^- \setminus A^-$ such that $\forall x \in B^+ \setminus A^+ \cup A^- \setminus B^-, \pi(x^*) > \pi(x)$. If many elements satisfy this property, let $x^*$ be one of those maximizing $\pi$. So, at any level $\lambda >_\pi (x^*)$, $A^+ \setminus B^+ \cup B^- \setminus A^-$ is empty; thus for these levels, $A^+_\lambda = B^+_\lambda$ and $A^-_\lambda = B^-_\lambda$, which imply the equality of the cardinalities. On the other hand, at level $\delta^* = \pi(x^*)$, there is no element in $B^+ \setminus A^+ \cup A^- \setminus B^-$. So, $|(B^+ \setminus A^+)_{\delta^*}| = 0$ and $|(A^- \setminus B^-)_{\delta^*}| = 0$. And there is at least one element in $A^+ \setminus B^+ \cup B^- \setminus A^-$ so $|(A^+ \setminus B^+)_{\delta^*}| \geq 1$ or $|(B^- \setminus A^-)_{\delta^*}| \geq 1$ (or even both).

Suppose now $|(A^+ \setminus B^+)_{\delta^*}| \geq 1$: from $|(B^+ \setminus A^+)_{\delta^*}| = 0$ and adding the common elements, we get $|B_{\delta^*}^+| = |(A^+ \cap B^+)_{\delta^*}|$. Since $|(A^+ \setminus B^+)_{\delta^*}| \geq 1$ we get $|A_{\delta^*}^+||B_{\delta^*}^+|$. From $|(A^- \setminus B^-)_{\delta^*}| = 0$ and adding the common element we get $|A^-| = |B^- \cap A^-|$ thus $|A^-| \leq |B^-|$. So $A \succ^{\text{BiLexi}} B$. We get $A \succ^{\text{BiLexi}} B$ in the same way from $|(B^- \setminus A^-)_{\delta^*}| \geq 1$. Hence $\succeq^{\text{BiLexi}}$ refines $\succeq^{\text{Discri}}$.

- Finally, suppose that $A \succ^{\text{BiLexi}} B$, i.e., at any level $\lambda$ higher than $\delta^*$, $|A_\lambda^+| = |B_\lambda^+|$ and $|A_\lambda^-| = |B_\lambda^-|$ and that at level $\delta^*$, there is a difference in favor of $A$. Then necessarily, at any level higher $\lambda$ than $\delta^*$, $|A_\lambda^+| - |A_\lambda^-| = |B_\lambda^+| - |B_\lambda^-|$. Let first suppose that at level $\delta^*$, the difference is made on the positive scale, i.e., $|A_{\delta^*}^+| > |B_{\delta^*}^+|$ and $|A_{\delta^*}^-| \leq |B_{\delta^*}^+|$. Summing up the inequalities we get: $|A_{\delta^*}^+| - |A_{\delta^*}^-||B_{\delta^*}^+| - |B_{\delta^*}^-|$. Then $A \succ^{\text{Lexi}} B$. If the difference is rather made on the negative side, we get $A \succ^{\text{Lexi}} B$ in a similar way. So, $A \succ^{\text{Lexi}} B$. Hence $\succeq^{\text{Lexi}}$ refines $\succeq^{\text{BiLexi}}$.

$\square$

*Proposition 10* . From $C \sim D$ and $A \cap (C \cup D) = \emptyset$, preferential independence implies $A \cup C \sim A \cup D$. Then, by transitivity: $A \cup C \succeq B$ implies $A \cup D \succeq B$; $A \cup D \succeq B$ implies $A \cup C \succeq B$; $B \succeq A \cup C$ implies $B \succeq A \cup D$; and $B \succeq A \cup D$ implies $B \succeq A \cup C$. $\square$

*Corollary 1* . Since $C \sim \emptyset$ and $A \cap C = \emptyset$, we can apply the principle of preferential independence and get $A \cup C \sim A$. Then, by transitivity, $A \succeq B$ implies $A \cup C \succeq B$. Conversely, $A \cup C \succeq B$ and $A \cup C \sim A$ implies, also by transitivity, that $A \succeq B$. Similarly, from $B \succeq A$ and $A \cup C \sim A$, transitivity implies $B \succeq A \cup C$; from $B \succeq A \cup C$ and $A \cup C \sim A$, transitivity implies $B \succeq A$. $\square$

*Corollary 2* . From the axiom of preferential independence, and because $(A \cup B) \cap C = \emptyset$: $A \succeq B \iff A \cup C \succeq B \cup C$. Then applying Proposition 10 and because $B \cap (C \cup D) = \emptyset$, $A \succeq B \iff A \cup C \succeq B \cup D$. $\square$

*Corollary 3.*

**Case $x \in B$:** Then by Proposition 10 we can replace $x$ by $y$ in $B$ and get $A \succeq B \iff A \succeq (B \setminus \{x\}) \cup \{y\}$. Let us then apply preferential independence and add $x$ on both sides of $\succeq$: $A \succeq (B \setminus \{x\}) \cup \{y\} \iff A \cup \{x\} \succeq (B \setminus \{x\}) \cup \{y\} \cup \{x\}$, i.e., $A \succeq (B \setminus \{x\}) \cup \{y\} \iff A \cup \{x\} \succeq B \cup \{y\}$, thus $A \succeq B \iff A \cup \{x\} \succeq B \cup \{y\}$.

**Case $x \notin B$:** Then preferential independence means that $A \succeq B \iff A \cup \{x\} \succeq B \cup \{x\}$ and Proposition 10 is used to get $A \succeq B \iff A \cup \{x\} \succeq B \cup \{y\}$.

$\square$

*Theorem 2* . It is easy to show that item 2 implies item 1. We have already seen that $\succeq^{\text{Lexi}}$ is complete, transitive, satisfies preferential independence, and refines $\succeq^{\text{BiPoss}}$. It is also easy to show that the grounding relations on $X$ induced by both relations are equivalent, i.e., that $\succeq^{\text{Lexi}}$ is an unbiased refinement of $\succeq^{\text{BiPoss}}$.

Let us now prove that item 1 implies item 2. Let $\succeq$ be a complete and transitive monotonic bipolar set-relation that satisfies preferential independence, and is an unbiased refinement of $\succeq^{\text{BiPoss}}$. Since $\succeq$ is an unbiased refinement of $\succeq^{\text{BiPoss}}$, $\succeq_{\underline{\mathbb{X}}} \equiv \succeq_{\underline{\mathbb{X}}}^{\text{BiPoss}}$. It is equivalently defined by $\pi$. The notions of $\lambda$-section, positive $\lambda$-section and negative $\lambda$-section are thus well defined.

1. Let us first suppose that, for any $\gamma$, $|A_\gamma^+| - |A_\gamma^-| = |B_\gamma^+| - |B_\gamma^-|$. Then we show that $\forall \gamma, A_\gamma \sim B_\gamma$. Two cases are possible:

   **Case** $|A_\gamma^+| - |A_\gamma^-| \geq 0$: Then it is possible to partition $A_\gamma$ into $n^A = |A_\gamma^-|$ pairs of $\{x^+, x^-\}$ and $n = |A_\gamma^+| - |A_\gamma^-|$ singletons $\{x^+\}$. Similarly, it is possible to partition $B_\gamma$ into $n^B = |B_\gamma^-|$ pairs of $\{y^+, y^-\}$ and $n$ singletons $\{y^+\}$—the same $n$ as for $A$, since $|A_\gamma^+| - |A_\gamma^-| = |B_\gamma^+| - |B_\gamma^-|$. So, from $\emptyset = \emptyset$, we can use Proposition 3 to add the $n$ singletons $\{x^+\}$ and the $n$ singletons $\{y^+\}$ on each side of the equivalence (left side for the $x^+$, right side for the $y^+$). Then Corollary 1 allows us to add the pairs $\{x^+, x^-\}$ on the left side, the pairs $\{y^+, y^-\}$ on the right side. Then we arrive at $A_\gamma \sim B_\gamma$.

   **Case** $|A_\gamma^+| - |A_\gamma^-| \leq 0$: The proof is similar. It is possible to partition $A_\gamma$ into $n^A = |A_\gamma^-|$ pairs of $\{x^+, x^-\}$ and $n = |A_\gamma^+| - |A_\gamma^-|$ singletons $\{x^-\}$. Similarly, it is possible to partition $B_\gamma$ into $n^B = |B_\gamma^-|$ pairs of $\{y^+, y^-\}$ and $n$ singletons $\{y^-\}$. So, from $\emptyset = \emptyset$, we can use Proposition 3 and Corollary 1 allow to add the pairs $\{x^+, x^-\}$ and the remaining singletons $\{x^-\}$ on the left side, the pairs $\{y^+, y^-\}$ and the remaining singletons $\{y^-\}$ on the right side. We arrive at $A_\gamma \sim B_\gamma$.

   The $\lambda$-sections of $A \cup B$ are pairwise disjoint. Thanks to Corollary 2 we can then "sum" the equivalences $A_\gamma \sim B_\gamma$ and get $\bigcup_\gamma A_\gamma \sim \bigcup_\gamma B_\gamma$, i.e., $A \sim B$.

2. Let us now suppose that there is a $\lambda > 0_L$ such that (i) for any $\gamma > \lambda$, $|A_\gamma^+| - |A_\gamma^-| = |B_\gamma^+| - |B_\gamma^-|$ and (ii) $|A_\lambda^+| - |A_\lambda^-| > |B_\lambda^+| - |B_\lambda^-|$. From a proof similar to the one just before, we get $\bigcup_{\gamma > \lambda} A_\gamma \sim \bigcup_{\gamma > \lambda} B_\gamma$. Three cases are then possible:

   **Case** $|A_\gamma^+| - |A_\gamma^-| \geq 0$: Then it is possible to partition $A_\gamma$ into $n^A = |A_\gamma^-|$ pairs of $\{x^+, x^-\}$ and $n = |A_\gamma^+| - |A_\gamma^-|$ singletons $\{x^+\}$. Similarly, it is possible to partition $B_\gamma$ into $n^B = \min(|B_\gamma^+|, |B_\gamma^-|)$ pairs of $\{y^+, y^-\}$ and $m$ singletons $\{y^+\}$ and a number $m'$ of singletons $\{y^-\}$, with $\max(m, m') = 0$. Because $|B_\gamma^+| - |B_\gamma^-| < |A_\gamma^+| - |A_\gamma^-| = n$, $m < n$. Let $x^*$ be one of the $n$ positive singletons in the partition of $A_\lambda$. Obviously, $x^* \cup_{\gamma < \lambda} A_\gamma^- \succ^{\text{BiPoss}} \cup_{\gamma < \lambda} B_\gamma^+$ and thus by monotony $x^* \cup_{\gamma < \lambda} A_\gamma \succ^{\text{BiPoss}} \cup_{\gamma < \lambda} B_\gamma$. Then $x^* \cup_{\gamma < \lambda} A_\gamma \succ \cup_{\gamma < \lambda} B_\gamma$ (this is the refinement hypothesis). Then we can add the $n$ singletons $x^+$ and $n$ singleton $y^+$ on the left and right side of the strict preference without modifying it (this is Proposition 3). By monotony, we can then add the singletons $y^-$, if any, and the remaining $m - n - 1$ singletons $x^+$, if any. Since the pairs $\{x^+, x^-\}$ (resp., $\{y^+, y^-\}$) are equivalent to $\emptyset$, they can be added on the left (resp., right) side without modifying the inequality (we use Corollary 1). We get $\cup_{\gamma \leq \lambda} A_\gamma \succ \cup_{\gamma \leq \lambda} B_\gamma$.

   **Case** $|B_\gamma^+| - |B_\gamma^-| \leq 0$: The proof if very similar to the previous one, from a negative $y^*$. We partition $B_\lambda$ into a maximum number of pairs, and a given number $n$ of negative singletons. $A_\lambda$ is also partitioned into a maximum number of pairs, possibly $m < n$ negative singleton or (the two conditions are exclusive) some positive singletons we show that $\cup_{\gamma<\lambda}A_\gamma \succ \cup_{\gamma<\lambda}B_\gamma \cup \{y^*\}$, then add $m$ the negative singletons $x^-$ and the negative singletons $y^-$ on their respective sides, add the remaining negative $y^-$ or the remaining positive $x^+$. The pairs $\{x^+, x^-\}$ (resp., $\{y^+, y^-\}$) being equivalent to $\emptyset$, they can be added on the left (resp., right) side without modifying the inequality. We get $\cup_{\gamma\leq\lambda}A_\gamma \succ \cup_{\gamma\leq\lambda}B_\gamma$.

**Finally,** $|B_\gamma^+| - |B_\gamma^-| > 0$ and $|A_\gamma^+| - |A_\gamma^-| < 0$ is incompatible with the fact that $|A_\gamma^+| - |A_\gamma^-| > |B_\gamma^+| - |B_\gamma^-|$.

We thus get $\cup_{\gamma\leq\lambda}A_\gamma \succ \cup_{\gamma\leq\lambda}B_\gamma$. The $\lambda$-sections of $A \cup B$ are pairwise disjoint. Thanks to Corollary 2 we can then "sum" the equivalences $A_\gamma \sim B_\gamma$, $\lambda > \gamma$ to the string preference and get $\bigcup_\gamma A_\gamma \succ \bigcup_\gamma B_\gamma$, i.e. $A \succ B$.

3. We have shown that $A \succ^{lexi} B \implies A \succ B$ and $A \sim^{lexi} B \implies A \sim B$. Because the relations are complete, this means that they are equivalent: $A \succeq^{lexi} B \iff A \succeq B$.

$\square$